%% file: main.tex
\documentclass{article}

 \usepackage[main, final]{neurips_2026}

\usepackage[utf8]{inputenc} 
\usepackage[T1]{fontenc}    
\usepackage{hyperref}       
\usepackage{url}            
\usepackage{booktabs}       
\usepackage{amsfonts}       
\usepackage{nicefrac}       
\usepackage{microtype}      
\usepackage{xcolor}         
\usepackage{enumitem}
\usepackage{amsmath}
\usepackage{longtable}
\usepackage{booktabs}  
\usepackage{array, makecell}
\usepackage{xcolor}    
\usepackage{pifont}   
\usepackage{amsmath}
\usepackage{amssymb}
\usepackage{mathtools}
\usepackage{amsthm}
\usepackage{subfiles}
\usepackage{hyperref}
\usepackage{url}
\usepackage{subfiles}
\usepackage{enumitem}
\usepackage{multirow}
\usepackage{graphicx}
\usepackage{booktabs}
\usepackage{color, colortbl}
\usepackage{multicol}
\usepackage{multirow}
\usepackage{amsmath} 
\usepackage{amssymb} 
\usepackage{wrapfig} 
\usepackage{kotex}
\usepackage{color, colortbl,enumitem}
\usepackage{tabularx}
\usepackage{caption}
\usepackage{titletoc}
\usepackage{adjustbox}
\usepackage{algorithm}
\usepackage{algpseudocode}
\usepackage{amsfonts}
\usepackage{bm}
\usepackage{latexsym}
\usepackage{tikz-cd}
\usepackage{amscd}
\usepackage{enumitem}
\usepackage{wrapfig}
\definecolor{lightgrayline}{gray}{0.85}
\newcommand{\cmark}{\textcolor{green}{\ding{51}}}  
\newcommand{\xmark}{\textcolor{red}{\ding{55}}}    

\hypersetup{linktocpage=true}
\newcommand\DoToC{%
    \startcontents
    \printcontents{}{1}{\vskip3pt\hrule\vskip5pt}
    \vskip3pt\hrule\vskip5pt
}
\newcommand{\proposed}{\textsc{PAIR}}
\title{PAIR: Prefix-Aware Internal Reward Model for Multi-Turn Agent Optimization}

\author{%
  Wonjoong Kim$^1$, Yeonjun In$^1$, Sangwu Park$^1$, Dongha Lee$^2$, Chanyoung Park$^1$\thanks{Corresponding Author} \\
  $^1$KAIST \hspace{2ex} $^2$Yonsei University\\
  \texttt{\{wjkim, yeonjun.in, sangwu.park, cy.park\}@kaist.ac.kr} \\
  \texttt{donalee@yonsei.ac.kr}
}

\begin{document}

\maketitle

\input{sections/000_abstract}
\input{sections/010_introduction}
\input{sections/020_related_work}
\input{sections/030_pair}

\input{sections/040_grpo}

\input{sections/060_conclusion}

\clearpage
\bibliographystyle{plain}
\bibliography{bibliography}



\clearpage
\appendix

\rule[0pt]{\columnwidth}{4pt}
\begin{center}
    \Large{\bf{Supplementary Material for \\[0.5em]
    PAIR: Prefix-Aware Internal Reward Model for \\[0.5em]
   Multi-Turn Agent Optimization}}
\end{center}
\vspace{2ex}

\DoToC

\clearpage
\input{sections/090_appendix}


\clearpage
\input{checklist.tex}

\end{document}

%% file: sections/000_abstract.tex
\begin{abstract}

A significant hurdle for current LLMs is the execution of complex, multi-stage tasks. 
Group Relative Policy Optimization (GRPO) has been emerging as a leading choice, but its reliance on sparse outcome rewards severely limits credit assignment across intermediate steps. Existing remedies—running full rollouts to assign step-level advantages, calling external LLM judges at each step, or computing intrinsic rewards that require ground-truth answers at every evaluation—introduce significant costs or practical constraints. 
We hypothesize that internal correctness probing over LLM hidden states can be repurposed as a step-level reward signal, potentially addressing all of these limitations at once. 
However, existing probing research assumes clean inputs, and we first show that this assumption breaks down in multi-step settings: hidden-state probes degrade severely under prefix contamination---tracking coherence with the (possibly corrupted) prefix rather than grounded correctness---while attention-based features remain robust to contamination but underperform on clean prefixes.
Building on this complementary relationship, we propose the Prefix-Aware Internal Reward (\proposed)---a two-stage model with a frozen hidden-state probe estimating belief-consistency and a lightweight attention-based head correcting it toward grounded correctness. Experimental results show that \proposed~achieves the highest AUROC on contaminated trajectories while operating at negligible inference cost, enabling dense step-level reward signals for GRPO training without external model calls, ground-truth dependencies, or full-trajectory rollouts\footnote{Our source code is available at \url{https://github.com/wonjoong-kim/PAIR}.}.

\end{abstract}

%% file: sections/010_introduction.tex
\section{Introduction}\label{sec:intro}

Multi-step reasoning tasks—where an agent must handle a sequence of actions (e.g., API calls, tool invocations) and interpret environmental feedback before producing a final answer—have emerged as a central challenge for large language models~(LLMs). Training LLMs in such settings increasingly relies on reinforcement learning, and Group Relative Policy Optimization (GRPO) \citep{guo2025deepseek, shao2024deepseekmath} is widely adopted due to its simplicity and strong empirical performance. When applied to multi-step agent tasks, however, GRPO operates with only a \emph{sparse outcome reward}—a single scalar indicating whether the final answer is correct. This sparsity makes credit assignment across intermediate steps extremely difficult: the training signal cannot distinguish which steps contributed to success or failure, leading to slow convergence and poor sample efficiency. To mitigate this issue, several lines of work have attempted to introduce denser, step-level reward signals for multi-step agent training, but each comes with notable limitations.

One natural approach for step-level reward signals is to run the model to completion and then propagate the \textbf{outcome reward} backward to assign per-step advantages \citep{ji2025tree, zong20262}.
However, even when early steps are clearly erroneous, the agent cannot know this at generation time and must complete the full trajectory before any signal is available.
This produces a large volume of wasteful rollouts from already-failed trajectories, resulting in significant computational inefficiency.
An alternative is to query a strong \textbf{external LLM} at every step to evaluate that step's quality.
However, judging intermediate steps is inherently difficult, leading to na\"ive reward criteria (e.g., format validity) and the per-step API cost and latency make this approach impractical for online RL training at scale.
Other research leverage the model's own log-probabilities as a reward signal, but these approaches require the \textbf{ground-truth answer} at every reward computation: they calculate teacher-forced log-probabilities of the gold response conditioned on the current trajectory prefix \citep{wang2025information, xie2026tips}.
While training datasets may provide ground-truth answers, the key limitation is that this computation must occur at runtime for every rollout---an operational dependency that restricts applicability and adds non-trivial forward-pass cost.
Recent multi-turn RL studies~\citep{wang2025spa, xi2025agentprm, zhou2025sweet} 
address step-level credit assignment but universally rely on \textbf{external reward models}---separately trained networks or LLM judges---rather than leveraging the agent's own internal representations, incurring significant overhead in training and maintaining these external components. 

\medskip
\noindent
As illustrated by the comparison in Table \ref{tab:baseline_comparison}, there remains a significant gap in achieving a reward signal that is both dense and entirely independent from external models. These limitations raise a natural question: \emph{can we provide dense step-level rewards at every step, without full-trajectory rollouts, without external LLM calls, without ground-truth dependencies at runtime, and without a separate reward model---using only the agent's own internal states?}
\medskip


We propose extracting inherently encoded correctness information from an LLM's internal signals to serve as a dense, step-level reward for GRPO training. This approach is motivated by recent probing research showing that two complementary types of internal signals during generation---\textit{hidden states} \citep{wang2024latent} and \textit{attention patterns} \citep{chuang2024lookback, ostmeier2026attention}---implicitly carry information about model's own correctness.
It resolves all four limitations above simultaneously: no full rollouts needed (rewards are computed per step), no external LLM calls (probes run on the agent's own representations), no runtime ground-truth dependency (probes are trained offline and applied to new trajectories), and near-zero marginal cost. However, applying this idea to multi-step agent settings introduces a fundamental challenge: \emph{prefix contamination}. That is, during RL exploration, the agent inevitably generates erroneous intermediate steps, corrupting the prefix on which subsequent steps are conditioned. To assess whether internal probes remain reliable under such contamination, we conduct a systematic analysis of internal probe behavior under prefix contamination and uncover three findings:

\input{tables/compare_baselines}

\begin{itemize}[leftmargin=.1in]
    \item \textbf{Hidden-state probes degrade severely under contamination.}
    Hidden-state probes degrade severely under contamination because they encode coherence with the (possibly corrupted) prefix rather than grounded correctness, inverting the reward signal at repair turns.
    
    \item \textbf{Attention features are comparatively robust.}
    Attention features capture structural behavior and exhibit smaller drops under contamination.
    
    \item \textbf{The two signals are complementary.}
    The two signals are complementary: hidden-state dominates on clean prefixes, attention on contaminated ones, neither suffices alone.
\end{itemize}

Building on these findings, we propose the \textbf{P}refix-\textbf{A}ware \textbf{I}nternal \textbf{R}eward Model for Multi-Turn Agent Optimization (\textbf{\proposed}), a two-stage architecture that preserves the strength of hidden-state probes under clean conditions while correcting their failures under contamination. We get a belief-consistency score $s_{bc}$ from a pre-trained hidden-state probe. An attention-based correction head then takes the attention features together with $s_{bc}$ and predicts a final score $s_{final}$ that adjusts $s_{bc}$ toward grounded correctness, which serves as the step-level reward.
By this design, the model learns $s_{final} \approx s_{bc}$ under clean prefixes, where belief-consistency already aligns with grounded correctness, automatically preserving baseline performance.
Under contaminated prefixes, on the other hand, the attention-based head activates to correct the divergence.
It is worth noting that supervision is required only \emph{once, offline}; at inference time, the model operates at probe-level cost, making it directly usable as a dense step-level reward in GRPO training.

In summary, our main contributions are as follows:
\begin{itemize}[leftmargin=.1in]
    \item \textbf{Empirical:} We provide the first systematic study of internal correctness probe failure under prefix contamination in multi-step agent settings.
    \item \textbf{Mechanistic:} We demonstrate that hidden states track belief-consistency while attention tracks structural patterns, and that these two signals are complementary---a finding that explains why na\"ive probes fail and guides the design of robust alternatives.
    \item \textbf{Methodological:} We propose \proposed, which adaptively combines hidden-state and attention signals via a correction architecture, preserving clean-prefix performance while restoring robustness under contamination.
    \item \textbf{Practical:}  \proposed~operates without external LLM calls, without runtime ground-truth access, and at probe-level inference cost, enabling dense step-level rewards for GRPO training of multi-step reasoning agents.
\end{itemize}

%% file: tables/compare_baselines.tex

\begin{table}[t]
\caption{Comparison between reward mechanisms for GRPO based on their functional requirements.}
\label{tab:baseline_comparison}
\centering
\resizebox{1\linewidth}{!}{
\begin{tabular}{l | cccccc}
\cmidrule[1.5pt]{1-7}
& Outcome               & LLM-based \citep{wei2025reinforcing}                  & Tree-based \citep{ji2025tree, zong20262}           & Intrinsic \citep{wang2025information, xie2026tips}                 & PRM \citep{wang2025spa, xi2025agentprm, zhou2025sweet}               & \textbf{Ours}                  \\
\cmidrule[1.5pt]{1-7}
Dense reward              & \xmark          & \cmark & \cmark & \cmark & \cmark & \cmark \\
w.o. LLM           & \cmark & \xmark & \cmark  & \cmark & \cmark & \cmark \\
w.o. full rollout      & \xmark & \cmark  & \xmark  & \cmark & \cmark & \cmark \\
w.o. external func. & \cmark & \cmark & \cmark  & \cmark & \xmark & \cmark \\
w.o. ground-truth      & \xmark   & \xmark    & \xmark & \xmark & \xmark & \cmark \\
\cmidrule[1.5pt]{1-7}
\end{tabular}
\vspace{-2ex}
}
\end{table}

%% file: sections/020_related_work.tex
\section{Related Work}\label{sec:related}

\subsection{Reinforcement learning for LLM agents}

Reinforcement Learning from Human Feedback (RLHF) \citep{ouyang2022training} is the dominant paradigm for LLM alignment, originally instantiated with PPO \citep{schulman2017proximal, stiennon2020learning} but recently superseded by Group Relative Policy Optimization (GRPO)~\citep{guo2025deepseek, liu2025understanding, shao2024deepseekmath, zhang2025critique, zhou2026demystifying}, which removes PPO's value network by estimating the baseline from a group of rollouts per prompt. In multi-turn agentic settings, however, GRPO inherits a sparse outcome reward problem that limits credit assignment across intermediate steps.


\subsection{Step-level and process reward models}

To mitigate the limitations of sparse rewards, several studies have proposed dense, step-level feedback mechanisms. Process Reward Models (PRMs) provide supervision for intermediate reasoning steps rather than just the final answer \citep{lightman2023let, ma2023let,  wang2312math, wang2025visualprm, yin2025dynamic, zhang2024entropy, zhang2025lessons}. However, extending PRMs to agentic settings---where step correctness is context-dependent and not easily verified---remains challenging~\citep{xi2025agentprm}.

\noindent\textbf{Tree-based Rewards.} 
Methods such as AT$^2$PO \citep{zong20262} and Tree-GRPO \citep{ji2025tree} utilize tree-search and sibling rollout comparisons to derive step-level advantages. While effective, these methods incur high computational overhead due to the requirement of multiple rollouts per step.

\noindent\textbf{LLM-as-a-Judge.} 
Recent papers try to use an LLM as a judge for correctness, reducing the human cost \citep{gu2024survey, kim2025beyond}. Approaches like MT-PPO \citep{wei2025reinforcing} employ secondary LLMs (e.g., GPT-4) for rule-based evaluation of the quality of each turn. However, the associated API costs and inference latency make them less practical for large-scale online RL training.

\noindent\textbf{Intrinsic and External Rewards.} 
Recent works such as AgentPRM \citep{xi2025agentprm} and SWEET-RL \citep{zhou2025sweet} have explored multi-turn agent optimization, but they depend on external reward models. Other intrinsic methods like IGPO \citep{wang2025information} and TIPS \citep{xie2026tips} leverage the log-probabilities of ground-truth answers as a reward signal, though they remain dependent on the availability of ground-truth trajectories.

In contrast to the above approaches, we propose to derive step-level rewards pruely from the agent's own internal representations, simultaneously addressing all the limitations they impose: no full-trajectory rollouts, no external LLM judges, no separately trained reward models, and no dependence on ground-truth answers during runtime.

\subsection{Internal state probing for correctness prediction}

A growing body of work demonstrates that LLM internal representations can be reliable indicators of model and correctness without external verification. Recent methods like CoE (Chain-of-Embedding) \citep{wang2024latent} explore training-free evaluation by observing changes in hidden state magnitudes and angles across layers.
Furthermore, the entropy of attention heads are demonstrated to predict the correctness of LLM response \citep{ostmeier2026attention}, and
Lookback ratio analysis~\citep{chuang2024lookback} also identifies that attention to previous context can signal hallucination.
Despite their efficiency, these internal probes have primarily been validated in \textit{clean-prefix} and \textit{single-turn} settings. Our work addresses a critical gap: the failure of naive internal probes under \textit{prefix contamination} in multi-turn scenarios. Our proposed \proposed~reconciles this by adaptively combining hidden states with robust attention features to provide a reliable reward signal even in contaminated contexts.

%% file: sections/030_pair.tex
\section{Proposed Method}\label{sec:method}

\subsection{Motivation}

Before presenting our method, we establish the empirical motivation: internal correctness probes fail systematically when the input prefix contains errors from earlier steps.

\subsubsection{Problem Setup}
We consider agentic tasks that require multi-step reasoning, repeated tool use, and environment interaction. Let $x \in \mathcal{X}$ be a task description sampled from a task distribution $\mathcal{T}$, and a task agent interacts with the environment to produce an execution trace $\tau$ and a final output $\hat{y}$. This process is denoted as:
\begin{equation}
    x \to \tau \to \hat{y}.
\end{equation}

The execution trace $\tau$ is defined as a sequence of states, actions, and observations:
\begin{equation}
    \tau = (t_1, a_1, o_1, \dots, t_T, a_T, o_T)
\end{equation}
where $t_t$ is the agent's thought (reasoning), $a_t$ is the action taken (tool call), and $o_t$ is the observation (tool output) received from the environment at step $t$. 
Each action $a_t$ is generated by a policy $\pi_\theta$ conditioned on the prefix $p_t = (x, t_1, a_1, o_1, \ldots, o_{t-1})$.

\paragraph{Prefix contamination.}
We say the prefix $p_t$ is \emph{contaminated} if one or more earlier thoughts $t_j$ ($j < t$) or actions $a_j$ ($j < t$) are erroneous---e.g., containing reasoning errors, tool misuse, observation misinterpretation, or reliance on wrong information.
This is the typical condition during RL exploration, where the policy is imperfect and generates incorrect intermediate steps.

\paragraph{Evaluation setup.}
We construct matched clean/contaminated test pairs: for each reasoning trajectory, two versions of the prefix preceding the evaluation turn---one clean (all prior steps correct) and one contaminated (one or more prior steps replaced with erroneous alternatives)---are produced, with the evaluation turn held identical across the two so that any difference in probe accuracy isolates the effect of the prefix. We use Qwen-2.5-7B-Instruct~\citep{yang2024qwen25} as our backbone and extract five feature types from the evaluation turn's forward pass: three \emph{hidden-state} features (Last-Token, Mean-Pooled, Multi-Layer) and two \emph{attention-based} features (Attention, Hidden+Attn). All probes are linear, trained on the clean training split and evaluated on both clean and contaminated splits; we report AUROC. Detailed feature definitions, attention statistics, the probing protocol, and the AUROC metric are described in Appendix~\ref{app:data_construction}.

\subsubsection{Key Findings}
\label{sec:findings}

\begin{wrapfigure}{r}{0.55\textwidth}
    \centering
    \includegraphics[width=\linewidth]{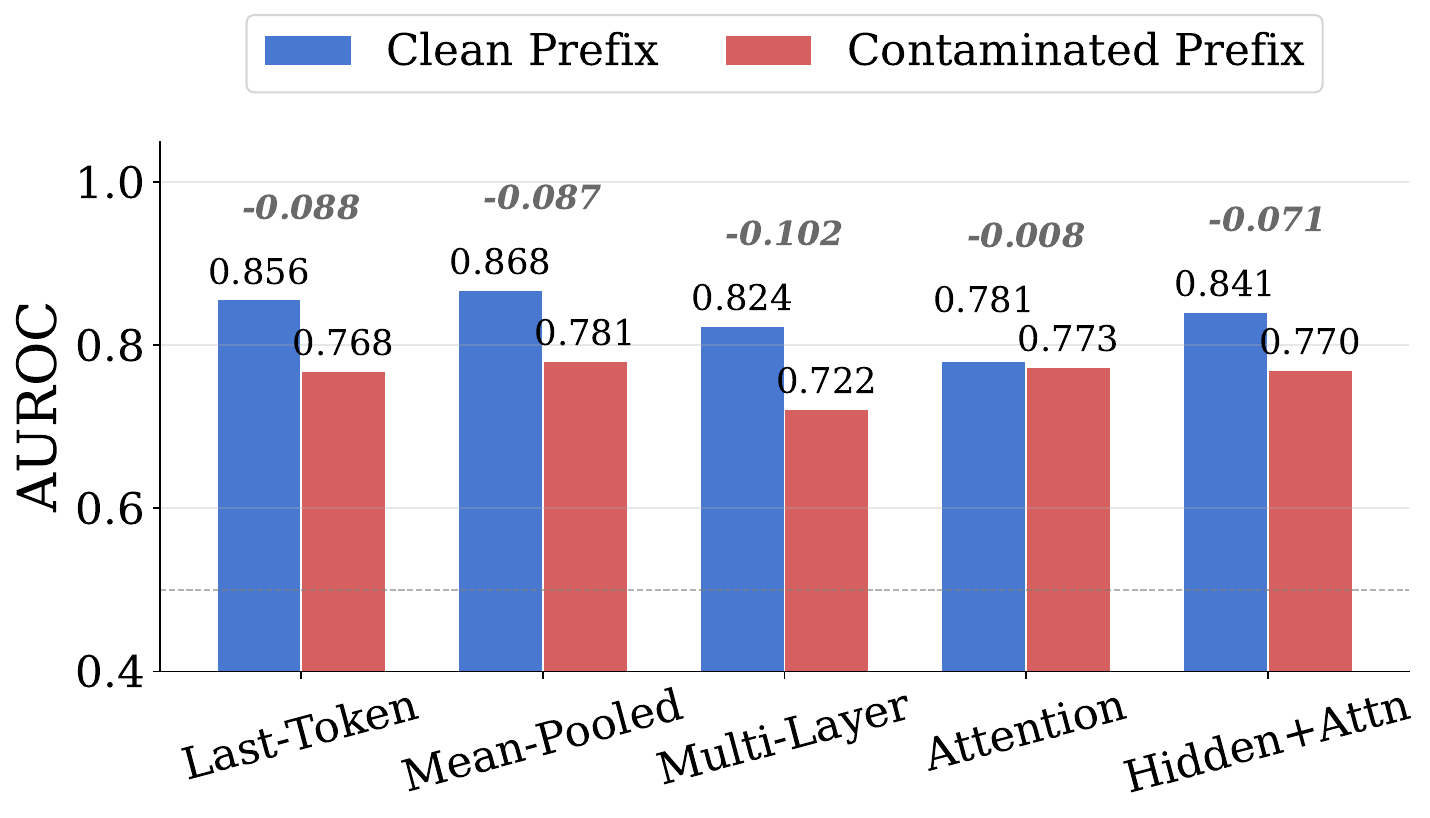}
    \caption{Comparison of AUROC degradation between hidden state probes and attention probes under clean vs. contaminated prefix settings.}
    \label{fig:combined_auroc_a}
    \vspace{-1ex}
\end{wrapfigure}
As shown in Figure \ref{fig:combined_auroc_a}, probes utilizing hidden-state exhibit significant AUROC drops when evaluated on contaminated prefixes compared to clean prefixes. However, attention probe shows smaller performance degradation under contamination.
While it underperforms hidden-state probes on clean data, its relative robustness under contamination makes them valuable in mixed conditions.
A na\"ive concatenation (Hidden+Attn) does not resolve this tension, as the hidden-state component can dominate and drag performance down under contamination.

This asymmetry is consistent with how LLM hidden states are shaped during pre-training: the next-token-prediction objective optimizes representations to capture \emph{what is coherent with the prefix}, not what is objectively correct for the underlying task—the model receives no direct supervision for the latter. This view also aligns with prior probing studies, which show that internal probes tend to recover the model's beliefs rather than external truth~\citep{burns2022discovering, kadavath2022language}.

Attention features, by contrast, capture \emph{structural} behavior patterns---e.g., whether the model attends to task-relevant tokens versus merely following the corrupted narrative---which remain informative regardless of prefix quality. This robustness is consistent with the architectural role of attention: rather than encoding accumulated content, attention determines \emph{where} information is routed at each step, and prior probing studies have successfully exploited this structural signal for hallucination detection and faithfulness analysis~\citep{chuang2024lookback, ostmeier2026attention}.

Based on these observations, we hypothesize that the hidden state encodes \emph{belief-consistency}---whether the current step is coherent with the accumulated prefix---rather than \emph{grounded correctness}---whether the current step makes genuine progress toward the task goal. 
In other words, these two notions align under clean prefixes: a step that is coherent with a correct history is also likely correct.
Under contaminated prefixes, on the other hand, they diverge: a step that perpetuates prior errors is belief-consistent but not grounded-correct, while a \emph{repair step} that corrects a prior error is grounded-correct but belief-inconsistent.

\subsubsection{Belief Consistency vs. Grounded Correctness}
The preceding experiments establish that hidden-state probes degrade under prefix contamination while attention probes remain relatively robust. This raises a deeper mechanistic question: \emph{do hidden-state probes track belief-consistency (coherence with the prefix) rather than grounded correctness (actual task utility)?} To answer this, we construct a controlled diagnostic dataset in which the two notions are deliberately placed in opposition.

\paragraph{Diagnostic construction.}
Each evaluation point is a triple (contaminated prefix, candidate turn, correctness label) in which coherence with the prefix and grounded correctness are deliberately anti-correlated. We construct two complementary turn types: \textbf{consistent-but-incorrect turns} that fluently follow the contaminated prefix and perpetuate the prior error (label \texttt{incorrect}), and \textbf{inconsistent-but-correct (repair) turns} that perform the objectively correct action even though they contradict the corrupted prefix (label \texttt{correct}). Unlike the previous experiment which mixed clean and contaminated prefixes, this diagnostic restricts to contaminated prefixes paired with strongly opposing turn types, isolating the belief-consistency / grounded-correctness effect. By construction, a probe tracking belief-consistency will rank consistent-but-incorrect above repair turns (AUROC at or below chance), while a probe tracking grounded correctness will rank correctly (AUROC well above chance); performance on this dataset is therefore a direct read-out of \emph{which signal a probe is tracking}.

\paragraph{Results.}
\begin{wrapfigure}{r}{0.5\textwidth}
    \centering
    \includegraphics[width=\linewidth]{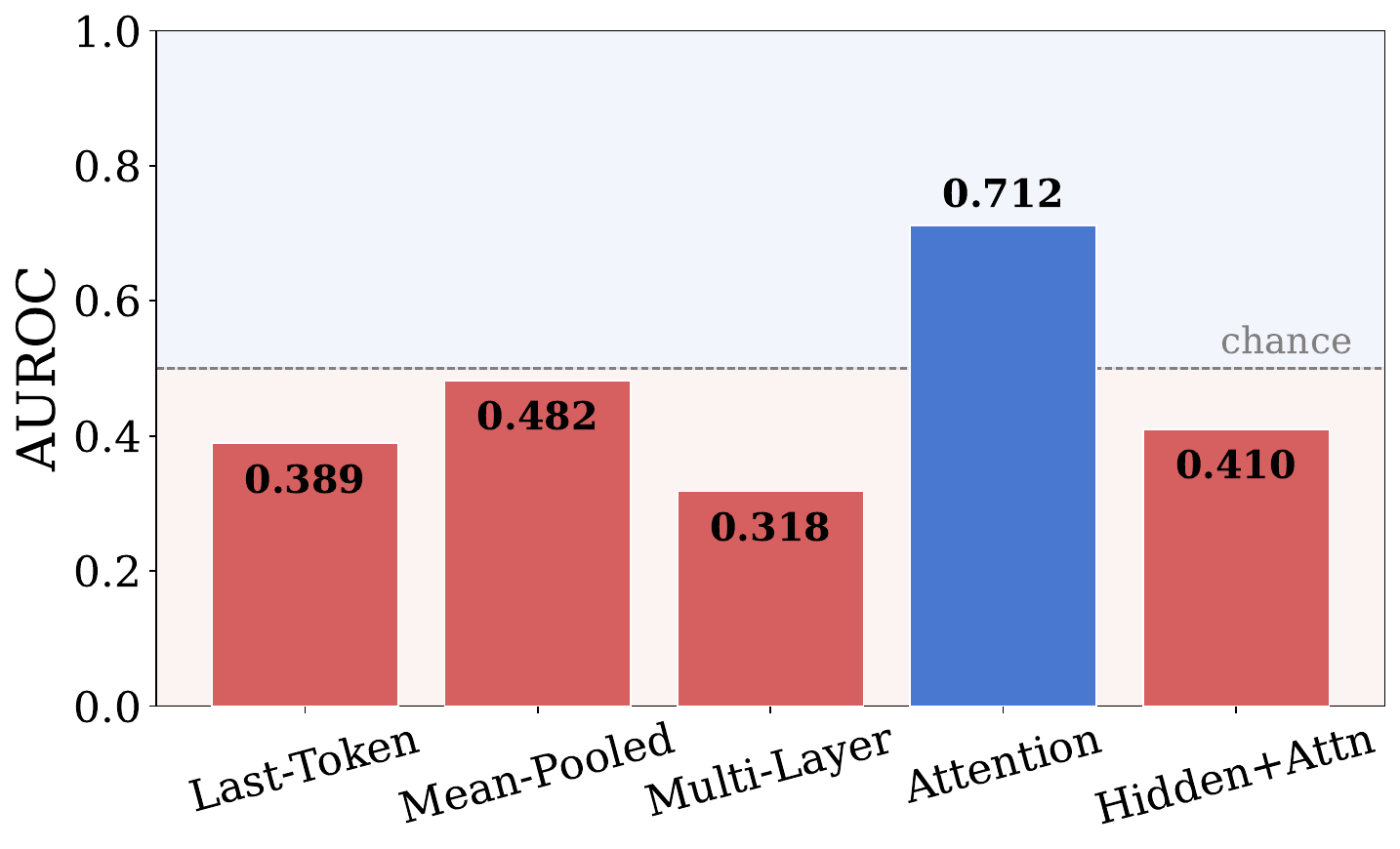}
    \caption{Comparison of hidden state and attention probes on the \textbf{adversarial diagnostic set}, where belief-consistency and grounded correctness are deliberately placed opposite.}
    \label{fig:combined_auroc_b}
    \vspace{-1ex}
\end{wrapfigure}
Figure~\ref{fig:combined_auroc_b} confirms the hypothesis. All three hidden-state probes score \emph{at or below chance}, meaning they consistently rank consistent-but-incorrect turns above repair turns---in an RL training loop, using such a probe as a reward would actively \emph{penalize} repair and \emph{reward} error-perpetuating behavior. The attention probe instead achieves AUROC well above chance, correctly identifying repair turns as more likely correct, indicating that attention patterns capture structural signals (whether the model attends to task-relevant information) that remain aligned with grounded correctness even under contamination. Naive concatenation (Hidden+Attn) does not resolve the conflict: the inverted hidden-state signal cancels the attention signal, yielding near-chance performance and underscoring the need for an \emph{adaptive} integration strategy. These findings provide direct mechanistic evidence for the design of \proposed: the hidden-state probe captures a meaningful but incomplete signal (belief-consistency), and the attention-based correction head compensates exactly when belief-consistency diverges from grounded correctness. Further sensitivity analysis regarding contamination distance is in Appendix~\ref{app:sensitivity}.

\subsection{Prefix-Aware Internal Reward Model (\proposed)}

Our empirical findings expose a \textbf{belief-consistency bias} in hidden-state probes: since they encode coherence with the prefix rather than grounded correctness, using them directly as a reward in multi-turn RL would reward steps that perpetuate prior errors and penalize \emph{repair turns} that contradict the contaminated history. Attention features, by contrast, remain aligned with grounded correctness even under contamination, but underperform hidden-state probes when the prefix is clean. A practical reward model must therefore preserve the strength of hidden-state probes on clean trajectories while exploiting attention features to correct the bias on contaminated ones.

To address this, we propose the \textbf{Prefix-Aware Internal Reward Model (PAIR)}. PAIR utilizes a two-stage architecture to decouple internal consistency from grounded utility:

\paragraph{Stage 1: Belief-consistency estimator.}
A pre-trained logistic regression probe maps hidden-state features to a belief-consistency score:
\begin{equation}
    s_{bc} = \sigma(\mathbf{w}_1^\top \mathbf{h} + b_1),
    \label{eq:stage1}
\end{equation}
where $\mathbf{h} \in \mathbb{R}^{d_{\text{model}}}$ is the last-token hidden state, $(\mathbf{w}_1, b_1)$ are the learned weights of the logistic regression probe, and $\sigma$ is the sigmoid function.
This probe is trained on clean data and then \emph{frozen}---it captures what standard probes do well.

\paragraph{Stage 2: Attention-based correction.}
A second logistic regression model takes the attention features concatenated with $s_{bc}$ and predicts the final correctness score:
\begin{equation}
    s_{final} = \sigma(\mathbf{w}_2^\top [\mathbf{a}; s_{bc}] + b_2),
    \label{eq:stage2}
\end{equation}
where $\mathbf{a} \in \mathbb{R}^{d_{\text{attn}}}$ is the multi-layer attention feature vector (4 statistics $\times$ $H$ heads $\times$ $L$ layers), and $(\mathbf{w}_2, b_2)$ are the learned weights of the correction head. The detailed ablation study of each features is presented in Appendix \ref{app:ablation_attn}. This correction head is trained on a mixture of clean and contaminated trajectories with the frozen Stage~1 score $s_{bc}$ as one of its inputs---it thereby learns to leave $s_{bc}$ alone on clean prefixes and to correct it where belief-consistency diverges from grounded correctness. Section~\ref{sec:unseen_contam} further shows that this correction transfers to contamination types excluded from Stage~2 training.

\textbf{Design rationale. }
Under clean prefixes, $s_{bc}$ already approximates grounded correctness accurately, so the optimal Stage~2 output is $s_{final} \approx s_{bc}$.
In contrast, the attention features detect structural anomalies under contaminated prefixes and produce $s_{final} \not\approx s_{bc} $ that corrects the belief-consistency estimate toward grounded correctness.
This design \emph{structurally} ensures that clean-prefix performance is preserved. The detailed training and inference process is presented in Appendix \ref{app:training}.

\subsection{Comparison with Baselines}
\label{sec:method_comparison}

\input{tables/result_pair}

We evaluate \proposed{} against a comprehensive set of internal probing baselines spanning three categories: 1) hidden state probes, 2) attention probes, and 3) unsupervised internal signal methods. 

\textbf{1) Hidden state probes.} \textit{Last-Token} trains a linear probe on the hidden state of the final generated token at the last layer.
\textit{Mean-Pooled} averages hidden states across all tokens in the generated response before probing.
\textit{Multi-Layer }concatenates hidden states from multiple layers of the last token, capturing representations at different levels of abstraction.

\textbf{2) Attention probes. }\textit{Attention} extracts summary statistics from attention weight of last layer—including maximum attention, standard deviation, prefix attention ratio, and self-attention ratio—and trains a linear probe on these features. \textit{Multi-Attn} concatenates same statistics with \textit{Attention} from all layers.
\textit{Hidden + Attn} concatenates the Multi-Layer hidden state features with the attention features and trains a single joint probe.

\textbf{3) Unsupervised internal signal methods.} \textit{Lookback Lens}~\citep{chuang2024lookback} detects hallucination by measuring the ratio of attention directed toward the context versus newly generated tokens.
\textit{Head Entropy}~\citep{ostmeier2026attention} uses the entropy of attention head distributions as an indicator of model uncertainty, where higher entropy suggests lower confidence in the generated content.
\textit{CoE-C}~\citep{wang2024latent} leverage Chain-of-Embedding trajectories from the hidden states on real-space and complex-space, respectively, to predict the correctness of the LLM responses.

\paragraph{Results.} Table~\ref{tab:detection_result} reports AUROC and Expected Callibration Error (ECE) on GTA and ToolBench. AUROC measures whether the probe ranks correct turns above incorrect ones, while ECE measures the gap between the probe's predicted probability and the empirical correctness frequency.
ECE matters in our setting because the probe's output is used as a continuous reward in GRPO, where overconfident or miscalibrated scores compress the within-group reward variance and weaken the resulting advantage signal.

\proposed{} achieves the highest AUROC across both datasets, outperforming all individual probes as well as unsupervised baselines. 
While \proposed{} does not achieve the lowest ECE in all cases, its calibration remains comparable to other probing methods and substantially better than the hidden state probes, which tend to produce overconfident predictions.
These results establish that \proposed{} provides a reliable correctness signal at the step level—a prerequisite for its deployment as a dense reward model in GRPO training.
Unlike LLM-as-judge approaches that incur additional inference costs per step, \proposed{} operates at probe-level computational overhead, making it practical for online RL settings where reward must be computed for every step of every rollout.

\subsection{Generalization to Unseen Contamination Types}
\label{sec:unseen_contam}

\input{tables/result_unseen_contam}

Since the Stage~2 correction head is trained on a mixture of clean and contaminated trajectories, one may ask whether it only works for the contamination types seen during training.
To test this, we run a leave-one-type-out study.
For each of the four contamination types (Appendix~\ref{app:data_construction}), we train Stage~2 on the other three types and measure AUROC on the held-out type.
We compare this with the full \proposed{}, which is trained on all four types and evaluated on the same held-out examples.
We also run a harder leave-two-types-out study, where Stage~2 is trained with only two types and evaluated on the remaining pair.

As shown in Table~\ref{tab:unseen_contam}, the AUROC drop stays within 0.025 for every held-out type.
In the leave-two-types-out setting, the drop stays within 0.032 for all six pairs, even though only half of the error types are seen during training.
This indicates that the correction head does not rely on type-specific error content.
Instead, it uses attention patterns shared across error types, which is consistent with our findings in Section~\ref{sec:findings}.
Note that the held-out sets for tool\_misuse and stale\_memory are small (22 and 10 negatives), so their estimates are less precise.

%% file: tables/result_pair.tex
\begin{table}[]
\caption{Step-level correctness prediction performance on GTA and ToolBench. Each evaluation point is a single (prefix, evaluation turn) pair labelled correct or incorrect, not a trajectory-level outcome. \textbf{AUROC} (higher is better) and \textbf{ECE} (lower is better) are reported for each method. All AUROC and ECE values are reported as mean over 3 random seeds (42, 43, 44).}
\label{tab:detection_result}
\centering
\resizebox{1\linewidth}{!}{
\begin{tabular}{l|l|>{\centering\arraybackslash}p{2.6cm}|>{\centering\arraybackslash}p{2.6cm}|>{\centering\arraybackslash}p{2.6cm}|>{\centering\arraybackslash}p{2.6cm}|>{\centering\arraybackslash}p{2.6cm}}
\cmidrule[1.5pt]{1-7}
\textbf{Dataset}                    & \textbf{Metric} & Last-token        & Mean-Pooled       & Multi-layer       & Attention         & Hidden + Attn                          \\ \cmidrule[1pt]{1-7}
\multirow{2}{*}{\textbf{GTA}}       & AUROC           & $0.8967_{\pm 0.012}$ & $0.8851_{\pm 0.014}$ & $0.9082_{\pm 0.010}$ & $0.8709_{\pm 0.015}$ & $0.8978_{\pm 0.013}$ \\
                                    & ECE             & $0.1294_{\pm 0.018}$ & $0.1521_{\pm 0.020}$ & $0.1342_{\pm 0.017}$ & $\mathbf{0.0954_{\pm 0.014}}$ & $0.1378_{\pm 0.018}$ \\ \cmidrule[0.5pt]{1-7}
\multirow{2}{*}{\textbf{ToolBench}} & AUROC           & $0.7654_{\pm 0.008}$ & $0.7372_{\pm 0.009}$ & $0.7821_{\pm 0.007}$ & $0.6847_{\pm 0.011}$ & $0.7598_{\pm 0.008}$ \\
                                    & ECE             & $0.2562_{\pm 0.014}$ & $0.2871_{\pm 0.016}$ & $0.2476_{\pm 0.013}$ & $0.1998_{\pm 0.011}$ & $0.2683_{\pm 0.015}$ \\ \cmidrule[1.5pt]{1-7}
\textbf{Dataset}                    & \textbf{Metric} & Multi-Attn        & Lookback Lens\citep{chuang2024lookback} & Head Entropy\citep{ostmeier2026attention} & CoE-C\citep{wang2024latent} & \textbf{Ours (\proposed)} \\ \cmidrule[1pt]{1-7}
\multirow{2}{*}{\textbf{GTA}}       & AUROC           & $0.9034_{\pm 0.011}$ & $0.8745_{\pm 0.014}$ & $0.8821_{\pm 0.013}$ & $0.7491_{\pm 0.022}$ & $\mathbf{0.9217_{\pm 0.009}}$ \\
                                    & ECE             & $0.1356_{\pm 0.018}$ & $0.1182_{\pm 0.015}$ & $0.1432_{\pm 0.019}$ & --                   & $0.1198_{\pm 0.014}$ \\ \cmidrule[0.5pt]{1-7}
\multirow{2}{*}{\textbf{ToolBench}} & AUROC           & $0.7765_{\pm 0.007}$ & $0.7295_{\pm 0.010}$ & $0.7561_{\pm 0.009}$ & $0.5953_{\pm 0.013}$ & $\mathbf{0.8129_{\pm 0.006}}$ \\
                                    & ECE             & $0.1961_{\pm 0.011}$ & $\mathbf{0.1685_{\pm 0.010}}$ & $0.2329_{\pm 0.013}$ & --                   & $0.2236_{\pm 0.012}$ \\ \cmidrule[1.5pt]{1-7}
\end{tabular}
\vspace{-2ex}
}
\end{table}

%% file: tables/result_unseen_contam.tex
\begin{table}[t]
\centering
\caption{AUROC of \proposed{} on contamination types excluded from Stage~2 training. Full \proposed{} is trained with all four types and evaluated on the same held-out examples. $\Delta$ = Full \proposed{} $-$ Held-out.}
\label{tab:unseen_contam}
\small
\begin{tabular}{lcccc}
\toprule
Held-out type(s) & \# Neg. & Held-out & Full \proposed{} & $\Delta$ \\
\midrule
\multicolumn{5}{l}{\textit{Leave-one-type-out}} \\
Reasoning error               & 54  & 0.9056 & 0.9259 & 0.020 \\
Observation misinterpretation & 52  & 0.9238 & 0.9311 & 0.007 \\
Tool misuse                   & 22  & 0.8508 & 0.8754 & 0.025 \\
Stale memory                  & 10  & 0.8742 & 0.8921 & 0.018 \\
\midrule
\multicolumn{5}{l}{\textit{Leave-two-types-out}} \\
Reasoning + Observation       & 106 & 0.9001 & 0.9285 & 0.028 \\
Reasoning + Tool misuse       & 76  & 0.8841 & 0.9113 & 0.027 \\
Reasoning + Stale memory      & 64  & 0.9025 & 0.9206 & 0.018 \\
Observation + Tool misuse     & 74  & 0.8860 & 0.9145 & 0.028 \\
Observation + Stale memory    & 62  & 0.9132 & 0.9248 & 0.012 \\
Tool misuse + Stale memory    & 32  & 0.8484 & 0.8806 & 0.032 \\
\bottomrule
\end{tabular}
\end{table}

%% file: sections/040_grpo.tex
\section{Downstream RL Experiments}
\label{sec:rl_experiments}


Having established that PAIR provides reliable step-level correctness signals even under prefix contamination (Section~\ref{sec:method}), we now evaluate whether these signals translate into effective reward models for multi-turn agent training via GRPO. 

\subsection{Experimental Setup}
\label{sec:rl_setup}

\textbf{Datasets.} We evaluate on two multi-step agent benchmarks that represent realistic agentic settings---as opposed to the search-augmented QA benchmarks (e.g., HotpotQA, NQ) used by prior internal reward methods.
In multi-step tool use tasks, there is no single gold answer string; multiple valid action sequences can achieve the same outcome, and errors propagate through environment state rather than merely resulting in missing information. Specifically,
\textbf{GTA} (General Tool Agents)~\citep{wang2024gta} is a benchmark containing 229 samples where agents use diverse real-world APIs.
Trajectories average 2--8 turns and require multi-step tool composition.
\textbf{ToolBench}~\citep{qin2023toolllm} is a dataset covering 16,000+ RapidAPI tools.
We stratified 1,000 samples (minimum 3 turns) from the full dataset.
Agents follow a Thought/Action/Action-Input/Observation loop. The detailed evaluation protocol is described in Appendix~\ref{app:eval}.

\textbf{Model.} We use \textbf{Qwen2.5-7B-Instruct}~\citep{yang2024qwen25} as the policy model for all GRPO experiments. Internal representations for PAIR are extracted from the same model being trained; no separate reward model is required. The detailed configuration for training is presented in Appendix \ref{app:grpo_config}.

\textbf{Reward baselines.} We compare PAIR against reward methods spanning five categories:

\begin{itemize}[leftmargin=1.5em]
    \item \textbf{Outcome-only}: Standard GRPO with a sparse binary reward assigned only at trajectory completion.
    \item \textbf{External reward}: \textit{LLM Judge} uses GPT-4o-mini to evaluate each step's quality on a 1--10 scale. Input prompt for LLM Judge is presented in Figure \ref{fig:llm_judge_prompt}.
    \item \textbf{Tree-based}: \textit{AT\textsuperscript{2}PO}~\citep{zong20262} performs tree-structured rollouts with entropy-weighted reward backpropagation. \textit{Tree-GRPO}~\citep{ji2025tree} uses intra- and inter-tree advantage normalization.
    \item \textbf{Internal probes}: We apply each probing method from Section~\ref{sec:method_comparison} directly as a step-level reward model, using the predicted correctness probability as the per-step reward. This includes hidden-state probes (\textit{Last-Token}, \textit{Mean-Pooled}, \textit{Multi-Layer}), attention probes (\textit{Attention}, \textit{Hidden+Attn}, \textit{Multi-Attn}), and unsupervised methods (\textit{Lookback Ratio}~\citep{chuang2024lookback}, \textit{Head Entropy}~\citep{ostmeier2026attention}, \textit{CoE}~\citep{wang2024latent}).
    \item \textbf{Intrinsic rewards}: \textit{IGPO}~\citep{wang2025information} computes information gain via teacher-forced gold log-probabilities. \textit{TIPS}~\citep{xie2026tips} uses potential-based reward shaping with a lagged teacher.
\end{itemize}

Our baselines fall into two structurally different classes. \textbf{(I) Reward sources for vanilla GRPO} share the same training pipeline and differ only in the per-step reward signal: \proposed, Outcome-only, LLM-as-a-judge, and all internal-probe baselines. \textbf{(II) Specialized RL training algorithms} change the optimization scheme itself (Tree-GRPO, AT$^2$PO, IGPO, and TIPS). Comparing across both classes tests whether \proposed---a reward source---can match methods that change the RL algorithm itself.

\textbf{\proposed~based Momentum Reward Integration.} 
At each step $t$ of a rollout, we extract the hidden state $\mathbf{h}_t$ and attention features $\mathbf{a}_t$ from the policy's forward pass and compute the \proposed~score $s_{final,t}$ via Eq.~\ref{eq:stage2}. Although $s_{final,t}$ can in principle be used directly as a step-level reward, doing so in GRPO leads to weak group-relative advantages: trajectories sharing a prompt receive similar absolute scores, so the within-group variance that GRPO relies on for credit assignment shrinks rapidly during training. We therefore use \proposed's score as the input to a \emph{momentum-based} reward that explicitly contrasts the current step against the trajectory's running mean, recovering the variance needed for advantage estimation. The detailed description of reward design is presented in Appendix \ref{app:reward}.

\subsection{Experimental Results}

\input{tables/main_result}

Table~\ref{tab:overall_result} reports success rate for all 16 methods, trained and evaluated separately on each of the two agent benchmarks (GTA and ToolBench). \proposed~achieves the best result on both GTA and ToolBench, outperforming the second-best baseline, LLM-as-a-judge. 

Two observations follow. First, PAIR is the only method that ranks at the top on both datasets, while every other approach is well-matched to one benchmark but mismatched to the other. Each baseline class has a fixed assumption that interacts unevenly with the structural differences between GTA and ToolBench: rollout-based methods (Tree-GRPO, AT$^2$PO) suffer noisy K-sample advantages on benchmarks with high branching; intrinsic methods (IGPO, TIPS) under-reward correct early steps in long trajectories; LLM-as-a-judge inherits prompt fit and surface bias for each task style; and single-signal probes commit to one feature type whose discriminative power varies with each benchmark's error distribution. PAIR alone has no such fixed assumption: it sources signal from the agent's own internals and adaptively combines two complementary signals.

Second, the only baseline that competes with \proposed~on either dataset is LLM-as-a-judge, and its competitiveness comes at a cost that is fundamentally incompatible with the use case we target. LLM-as-a-judge requires an external LLM call \emph{per step} of the agent's trajectory to obtain a reward signal: as a step-level reward source for online GRPO training, this incurs hundreds to thousands of additional API calls per training episode, multiplied by the number of episodes per training run. The resulting wall-clock latency and monetary cost make it effectively unusable as a dense step-level reward at any non-trivial training scale, regardless of its accuracy. \proposed, in contrast, runs a single forward pass of the agent's own model—reusing the activations the agent has already computed—and produces step-level rewards at probe-level marginal cost (logistic regression over a vector of pre-existing features). Further analysis for computing cost is presented in Appendix \ref{app:complexity}.


\paragraph{Effect of Momentum-based reward. }

\begin{wrapfigure}{r}{0.5\textwidth}
    \centering
    \vspace{-2ex}
    \includegraphics[width=\linewidth]{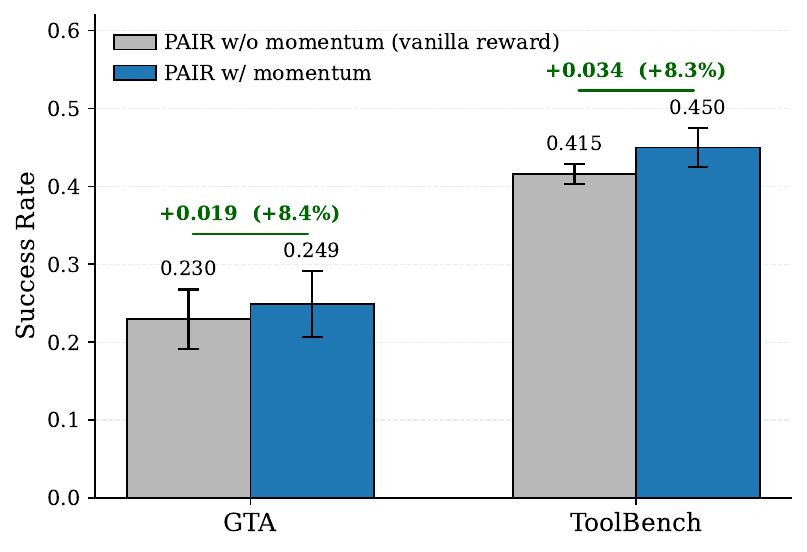}
    \caption{Effect of the momentum reward. \proposed~with the momentum-based reward outperforms the vanilla by $\approx 8\%$ on both GTA and ToolBench.}
    \label{fig:momentum_ablation}
    \vspace{-3ex}
\end{wrapfigure}
The probe output $s_{final,t}$ produced by \proposed's two-stage architecture can be used directly as a step-level reward in GRPO, without any further transformation. It already yields a competitive operating point: as Figure~\ref{fig:momentum_ablation} shows, the vanilla \proposed---which uses $s_{final,t}$ as the per-step reward attains $0.230$ on GTA and $0.415$ on ToolBench, on par with the strongest non-PAIR baselines in Table~\ref{tab:overall_result}. The probe alone is therefore already a usable replacement for external or rollout-based reward signals, validating the probe-as-reward design at the level of the underlying correctness predictor.
The momentum-based variant used in our main results goes one step further by adding an explicit contrast between each step's score and the trajectory's running mean. This reward-shaping step yields a consistent improvement on top of the vanilla variant. The takeaway is that \proposed's effectiveness does not rest solely on the probe being a good correctness predictor---\emph{how} the probe output is integrated into the GRPO objective also matters, and the momentum formulation extracts measurably more learning signal from the same underlying probe. More detailed hyperparameter analysis is described in Appendix \ref{app:hparam_analysis}.

\subsection{Cross-domain Transfer}

\begin{wrapfigure}{r}{0.55\textwidth}
    \centering
    \includegraphics[width=\linewidth]{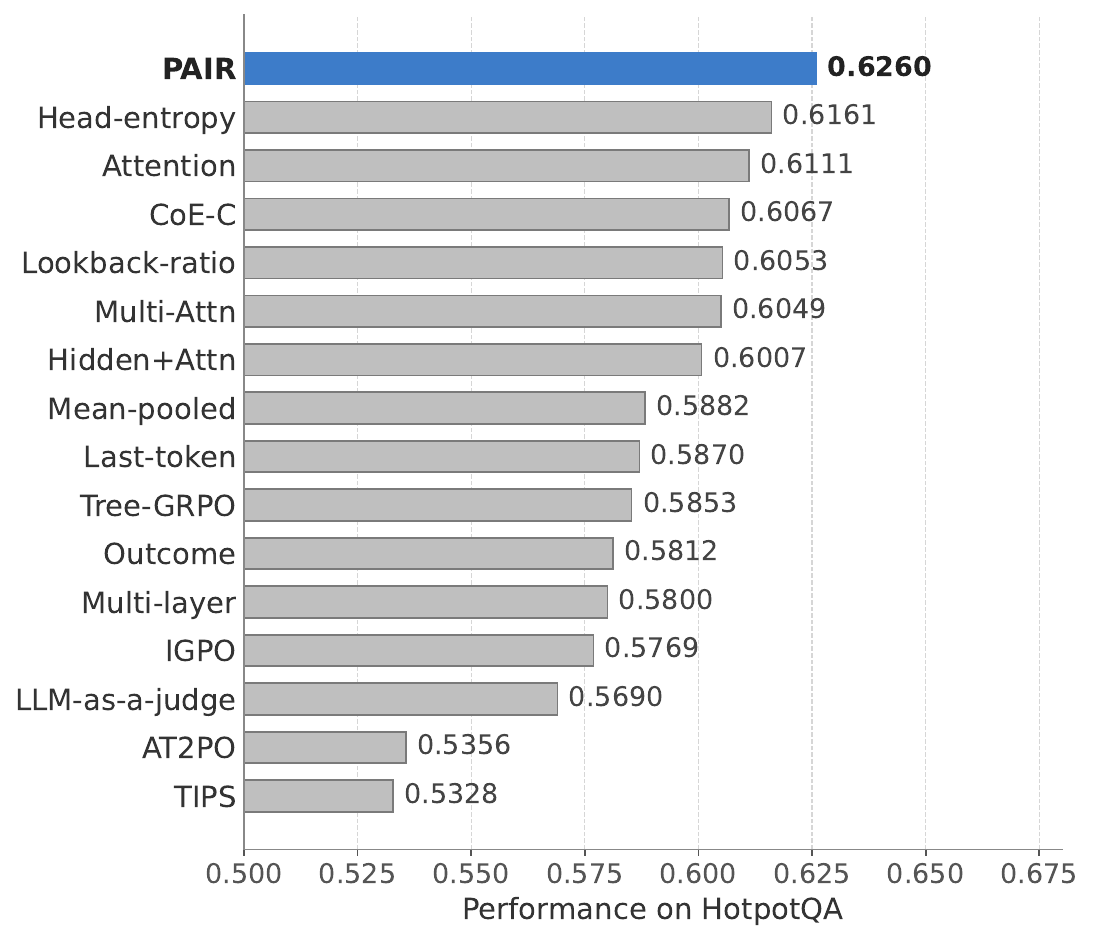}
    \caption{Transfer performance on HotpotQA. Each method is trained on GTA and evaluated on HotpotQA without further fine-tuning. \textbf{\proposed} achieves the best result among all 16 methods.}
    \label{fig:transfer_hotpotqa}
    \vspace{-1ex}
\end{wrapfigure}
To test whether \proposed's signal is anchored to a transferable property of the agent's computation rather than to dataset-specific surface features, we additionally evaluate cross-domain transfer: all reward models are trained on GTA and applied, without any adaptation, to HotpotQA—a multi-hop reasoning benchmark that shares the multi-step structure of agentic tasks but differs substantially in distribution. As shown in Figure~\ref{fig:transfer_hotpotqa}, \proposed~ranks first among all 16 methods, and three patterns in the ranking corroborate its design rationale.


\textbf{First}, attention-based methods (Head-entropy, Attention, Lookback-ratio) occupy the top of the table while every hidden-state-only probe falls to the middle. Attention patterns reflect structural aspects of the model's computation---properties of the underlying transformer mechanism rather than of task content---and therefore transfer more gracefully than hidden states, which encode semantic information of the training distribution. This extends the asymmetry observed under \emph{prefix contamination} (Section~\ref{sec:findings}) to \emph{distributional shift}.

\textbf{Second}, naive concatenation of hidden-state and attention features (Hidden+Attn) underperforms standalone Attention: the unreliable hidden-state component drags down the more robust attention signal under domain shift. \proposed~instead places hidden state in a fixed residual base and lets the attention head learn an additive correction in logit space, structurally protecting the attention signal—and ranks above both. The same two information sources thus yield a loss under concatenation but a gain under \proposed, isolating the value of the architectural decoupling itself.


\textbf{Third}, methods that incur rollout, external-model, or ground-truth costs occupy the bottom of the table: rollout-based methods (Tree-GRPO, AT$^2$PO) inherit high training-time variance from $K$-sample sibling estimates that compounds under domain shift; ground-truth-dependent methods (IGPO, TIPS) shape policies around GTA's gold-answer format, mismatching HotpotQA's short-span answers; and LLM-as-a-judge inherits judge biases tuned for GTA-style tasks while remaining prohibitive at training scale via API costs. Across all comparisons, \proposed~is the only method that simultaneously achieves the highest accuracy, operates without ground-truth at inference, requires no rollouts or external model calls, and transfers across domains without re-training---directly addressing the four limitations identified in the introduction.

%% file: tables/main_result.tex
\begin{table}[]
\caption{Overall performance comparison across all methods on GTA and ToolBench. Values are reported as mean$_{\pm\text{std}}$ over $3$ random seeds (42, 43, 44).}
\label{tab:overall_result}
\centering
\resizebox{0.85\linewidth}{!}{
\begin{tabular}{l >{\centering\arraybackslash}p{2.8cm} >{\centering\arraybackslash}p{2.8cm} >{\centering\arraybackslash}p{2.8cm} >{\centering\arraybackslash}p{2.8cm}}
\toprule
\textbf{Dataset}        & Outcome           & LLM-as-a-judge    & Last-token        & Multi-layer                                       \\
\midrule
\textbf{GTA}            & $0.1851_{\pm 0.055}$ & $\underline{0.2387_{\pm 0.042}}$ & $0.1372_{\pm 0.060}$ & $0.1789_{\pm 0.058}$ \\
\textbf{ToolBench}      & $0.3878_{\pm 0.035}$ & $\underline{0.4203_{\pm 0.028}}$ & $0.3837_{\pm 0.038}$ & $0.3856_{\pm 0.037}$ \\
\addlinespace[0.1em]
\midrule
\textbf{Dataset}        & Mean-pooled       & Attention         & Hidden+Attn       & Multi-Attn       \\
\midrule
\textbf{GTA}            & $0.1798_{\pm 0.060}$ & $0.2154_{\pm 0.048}$ & $0.1432_{\pm 0.062}$ & $0.2218_{\pm 0.046}$ \\
\textbf{ToolBench}      & $0.3917_{\pm 0.037}$ & $0.3471_{\pm 0.030}$ & $0.3784_{\pm 0.038}$ & $0.3614_{\pm 0.030}$ \\
\addlinespace[0.1em]
\midrule
\textbf{Dataset}        & Lookback-ratio \citep{chuang2024lookback} & Head-entropy \citep{ostmeier2026attention}  & CoE-C \citep{wang2024latent} & IGPO \citep{wang2025information} \\
\midrule
\textbf{GTA}            & $0.1812_{\pm 0.050}$ & $0.1989_{\pm 0.048}$ & $0.1721_{\pm 0.055}$ & $0.2005_{\pm 0.052}$ \\
\textbf{ToolBench}      & $0.3934_{\pm 0.030}$ & $0.3868_{\pm 0.030}$ & $0.4034_{\pm 0.034}$ & $0.3952_{\pm 0.032}$ \\
\addlinespace[0.1em]
\midrule
\textbf{Dataset}        & TIPS \citep{xie2026tips} & Tree-GRPO \citep{ji2025tree} & AT$^2$PO \citep{zong20262} & \textbf{\proposed}                       \\
\midrule
\textbf{GTA}            & $0.1856_{\pm 0.058}$ & $0.1934_{\pm 0.058}$ & $0.1865_{\pm 0.055}$ & $\mathbf{0.2489_{\pm 0.042}}$ \\
\textbf{ToolBench}      & $0.3483_{\pm 0.035}$ & $0.3935_{\pm 0.035}$ & $0.3812_{\pm 0.034}$ & $\mathbf{0.4498_{\pm 0.025}}$ \\
\bottomrule
\end{tabular}
\vspace{-3ex}
}
\end{table}

%% file: sections/060_conclusion.tex
\section{Conclusion}

We have identified a critical failure mode of internal correctness probes in multi-step agent settings: prefix contamination causes hidden-state probes to track belief-consistency rather than grounded correctness, producing reward signals that actively discourage error correction.
Through systematic analysis, we showed that attention features provide a complementary, more robust signal that captures structural behavior patterns.
Building on these findings, we proposed the \proposed, a two-stage reward architecture that freezes a hidden-state probe as a belief-consistency estimator and applies a lightweight attention-based correction to bridge the gap to grounded correctness.
\proposed~preserves strong performance under clean conditions while recovering robustness under contamination, all at probe-level inference cost.
By enabling dense, step-level reward signals without external model calls, runtime ground-truth dependencies, or separate reward model training, \proposed~makes probe-based reward practical for multi-step agent RL training.
We believe this work opens a promising direction at the intersection of internal state analysis and reinforcement learning for LLM agents.

%% file: sections/090_appendix.tex
\section{Data Contamination for Motivation Experience} \label{app:data_construction}

We present details of the construction of the matched clean / contaminated trajectory pairs used in the motivation experiments (Section~\ref{sec:findings}) and as the offline training corpus for \proposed~(Section~\ref{sec:rl_experiments}). We describe the source data (\S\ref{app:data_source}), the four contamination types and their dataset-specific adaptations (\S\ref{app:contam_types}), the construction of the diagnostic \emph{belief-consistency} and \emph{grounded-correctness} turns used in Figure~\ref{fig:combined_auroc_b} (\S\ref{app:diagnostic}), and the contamination procedure (\S\ref{app:contam_procedure}). The full prompts used to drive contamination with GPT-4o-mini are reproduced in Figure~\ref{fig:contamination_system_prompt} and \ref{fig:contamination_user_prompt}.

\subsection{Source Trajectories and Splits}\label{app:data_source}
 
We construct contaminated trajectories from two multi-turn agent benchmarks: \textbf{GTA}~\citep{wang2024gta} and \textbf{ToolBench}~\citep{qin2023toolllm}. Both follow the canonical agentic format
$$
\tau = (x, t_1, a_1, o_1, \dots, t_T, a_T, o_T),
$$
where $x$ is the task description, and each step consists of a thought $t_i$, an action / tool call $a_i$, and an observation $o_i$ returned by the environment. We apply two filters to obtain trajectories suitable for contamination: (i) at least \emph{three} assistant turns (so that contamination can be applied to a non-evaluation turn while leaving at least one upstream turn intact); and (ii) trajectory length $\le 8$ turns, which covers $> 96\%$ of episodes in both datasets and avoids context-window blow-up under our backbone (Qwen2.5-7B-Instruct, $4{,}096$-token context).
 
For each dataset we apply an $80{:}20$ stratified train / test split with a fixed seed ($42$). For ToolBench we additionally subsample $1{,}000$ episodes from the $48{,}454$-episode source release using stratified sampling on trajectory length, yielding the final episode counts in Table~\ref{tab:contam_stats}.

\input{tables/contam_stats}

\subsection{Contamination Types}\label{app:contam_types}
 
We instantiate four canonical failure modes observed in deployed agentic systems. Each type targets a specific component of the multi-step trajectory (reasoning, action, observation handling, or working memory), giving full coverage of the ways an upstream turn can damage downstream credit assignment.

\begin{itemize}[leftmargin=1.5em]
    \item \textbf{Reasoning error.} The chosen assistant thought $t_j$ is rewritten to contain a logically flawed inference that nonetheless appears fluent and contextually coherent. The corrupted thought is one that a reasonable but mistaken policy could plausibly produce.
    \item \textbf{Tool misuse.} The chosen action $a_j$ is rewritten to invoke an inappropriate tool, supply a malformed argument, or call the right tool with semantically wrong parameters.
    \item \textbf{Observation misinterpretation.} The chosen thought $t_j$ is rewritten so that it reaches a conclusion that is inconsistent with the immediately preceding observation $o_{j-1}$ — for example, claiming an API call succeeded when its response indicates failure, or extracting the wrong field from a tool output.
    \item \textbf{Stale memory.} The chosen turn is rewritten to rely on information from earlier in the trajectory in a way that contradicts later evidence — for example, persisting an outdated value as if it were current. 
\end{itemize}

\subsection{Diagnostic Pairs for the Belief-Consistency Hypothesis}\label{app:diagnostic}
 
The motivation experiment of Figure~\ref{fig:combined_auroc_b} requires turns where \emph{belief-consistency} and \emph{grounded correctness} are deliberately placed in opposition. We construct two complementary turn types for each base trajectory.
 
\paragraph{Belief-consistency turn (looks right, is wrong).} Given a \emph{clean} prefix, we generate a next assistant turn that is fluent and locally coherent with the conversation history but does not advance task completion: a redundant retrieval, an unnecessary clarification, or a syntactically valid tool call that retrieves task-irrelevant information. The turn is locally indistinguishable from a normal continuation but contributes nothing to the gold solution.
 
\paragraph{Grounded-correctness turn (looks wrong, is right).} Given a \emph{contaminated} prefix, we generate a next assistant turn that performs the objectively correct next action, regardless of whether it is locally coherent with the contaminated history. Concretely, the generator is conditioned on (i) the original \emph{clean} task description, (ii) the contaminated prefix, and (iii) the gold next action, and is instructed to produce the gold action even if it contradicts the immediately preceding (contaminated) turn. These are the analogue of \emph{repair turns}: turns that detect the prior error and override it.
 
The resulting maximally adversarial pair — a \emph{looks-right-but-wrong} turn after a clean prefix and a \emph{looks-wrong-but-right} turn after a contaminated prefix — is what the probes in Figure~\ref{fig:combined_auroc_b} are scored on.

\subsection{Contamination Procedure}\label{app:contam_procedure}
 
For each clean trajectory $\tau = (x, t_1, a_1, o_1, \dots, t_T, a_T, o_T)$ with $T \ge 3$ assistant turns, we designate the \emph{last} assistant turn $(t_T, a_T)$ as the \emph{evaluation turn} and the prefix $p_T = (x, t_1, a_1, o_1, \dots, o_{T-1})$ as the contamination target. The procedure is as follows:
 
\begin{enumerate}[leftmargin=1.5em]
\item \textbf{Pick a contamination target.} Sample one assistant turn $j \in \{1, \dots, T-1\}$ uniformly at random.
\item \textbf{Pick a contamination type.} Sample a contamination type from the set of types compatible with the chosen turn (e.g., \emph{tool\_misuse} requires that the chosen turn contain a tool call; cf.\ Table~\ref{tab:contam_compat}).
\item \textbf{Generate the corrupted turn.} Issue a single GPT-4o-mini call with the type-specific system prompt (Figure~\ref{fig:contamination_system_prompt} and Figure~\ref{fig:contamination_user_prompt}) and the relevant turn context, instructing the model to rewrite the chosen turn as a plausible-looking but objectively incorrect version of the original.
\item \textbf{Preserve downstream.} Replace turn $j$ in the trajectory with the corrupted version. \emph{Do not modify any other turn}, including all subsequent observations and the evaluation turn $(t_T, a_T)$. This preserves matched-pair semantics: each $(\tau^{\text{clean}}, \tau^{\text{contam}})$ pair differs only in turn $j$, and the evaluation turn over which we measure probe accuracy is identical across the two.
\item \textbf{Log metadata.} For every contaminated episode we record the contamination type, the index $j$ of the corrupted turn, the original turn content, and the GPT-4o-mini response, enabling downstream auditing and validation.
\end{enumerate}

\input{tables/contam_compatibility}
 
The \emph{evaluation turn is held identical} across the two versions of the trajectory. Any difference in probe AUROC between $\tau^{\text{clean}}$ and $\tau^{\text{contam}}$ can therefore be attributed to the prefix alone, not to differences in the step being scored. This matched-pair design is what makes the contamination effect causally identifiable in Figure~\ref{fig:combined_auroc_a}.

\subsection{Probing Protocol and Feature Definitions}\label{app:probing_protocol}

Each evaluation point in the test set is a triple (prefix, evaluation turn, correctness label), where the evaluation turn may be either correct or incorrect with respect to the gold task solution. Features are extracted by running the backbone forward through \texttt{[prefix; evaluation turn]} under teacher forcing---i.e., the evaluation turn is supplied as input rather than autoregressively generated---and reading out the hidden state and attention statistics at the evaluation turn's tokens. The probing task is a binary classification of the correctness label, and we report AUROC, the probability that the probe assigns a higher score to a correct evaluation turn than to an incorrect one.

We extract five feature types from the evaluation turn's forward pass:
\begin{itemize}[leftmargin=1.3em,style=unboxed]
\item \textbf{Last-Token}: the hidden state of the final token at the last layer ($d_{\text{model}}$ dimensions).
\item \textbf{Mean-Pooled}: the mean of all token hidden states within the evaluation turn at the last layer.
\item \textbf{Multi-Layer}: concatenation of the last-token hidden states from the final $4$ layers ($4 \times d_{\text{model}}$ dimensions).
\item \textbf{Attention}: for each attention head in the last layer, four statistics over the evaluation turn's attention distribution---peak magnitude (\texttt{max\_attn}), spread (\texttt{std\_attn}), proportion attending to prefix tokens (\texttt{prefix\_ratio}), and proportion attending to own-turn tokens (\texttt{self\_ratio})---yielding a $4 \times H$-dimensional vector where $H$ is the number of heads.
\item \textbf{Hidden+Attn}: concatenation of Last-Token and Attention features.
\end{itemize}

All probes are \texttt{StandardScaler} $+$ \texttt{LogisticRegression} ($\ell_2$ penalty), trained on the clean training split and evaluated on both the clean and contaminated test splits.

\section{Sensitivity Analysis Regarding to Contamination Distance}\label{app:sensitivity}

\begin{figure}[t]  
\centering
    \includegraphics[width=1\linewidth]{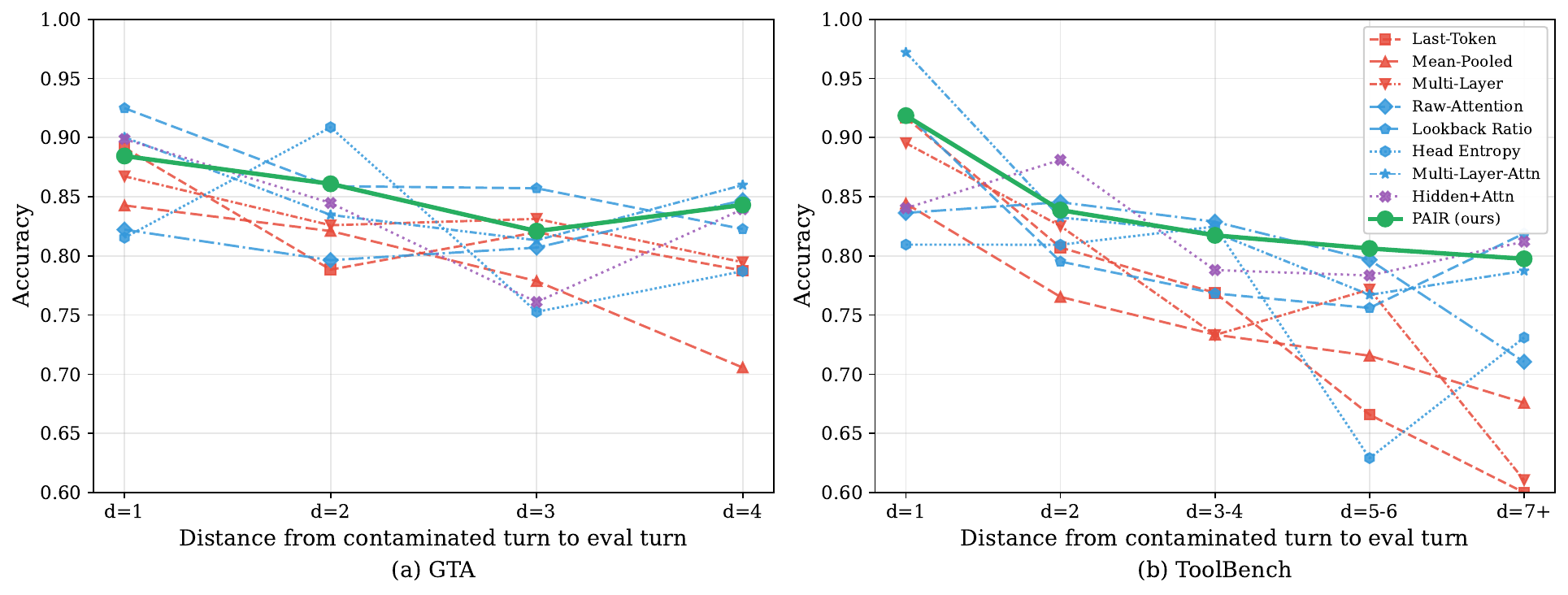} 
    \caption{Correctness prediction performance of linear probing models regarding to different contamination distances.}
    \label{fig:distance_sensitivity}
\end{figure}
We further examine how each probe's accuracy depends on the temporal distance $d$ between the contaminated turn and the evaluation turn ($d=1$: the error is the immediately preceding turn; larger $d$: the error sits deeper in the prefix). Figure~\ref{fig:distance_sensitivity} reports per-probe accuracy stratified by $d$ on both benchmarks.

Two patterns are visible across the two benchmarks. \textbf{First, all probes degrade as the contaminated turn moves further from the evaluation turn.} Long-range contaminations are harder to recover from the evaluation turn's local representations alone: by the time the evaluation turn is processed, the model's belief state has had several intervening (correct) turns to re-stabilize around the corrupted prefix, so the local features of the evaluation turn carry less direct evidence of the distant error. The effect is mild on GTA, whose trajectories are short ($d \le 4$), but is pronounced on ToolBench, where $d$ extends to $7+$ and accuracy spans roughly $0.60$ to $0.95$ across the range.

\textbf{Second, hidden-state probes degrade more steeply with distance than attention probes.} On ToolBench, the three hidden-state probes (Last-Token, Mean-Pooled, Multi-Layer) start from $\approx 0.85$--$0.90$ at $d=1$ and drop to $0.60$--$0.68$ at $d=7+$, whereas attention-based probes (Raw-Attention, Lookback Ratio, Head Entropy, Multi-Layer-Attn) decline more gently. This corroborates the mechanistic claim of Section~\ref{sec:findings}: hidden states encode \emph{prefix-conditional belief}, and the longer the prefix between the contamination and the evaluation turn, the more confidently the model's belief re-stabilizes around the corrupted narrative---producing hidden states that are \emph{more} coherent with the wrong prefix, not less. Attention features, by contrast, capture structural routing patterns that depend on \emph{how} the model is processing the current turn rather than on \emph{how long ago} the error occurred, and are therefore less sensitive to $d$.

\proposed~itself remains among the top performers across all distances and exhibits the smallest drop with $d$: on ToolBench it holds near $0.80$ accuracy even at $d=7+$, where every hidden-state probe has fallen below $0.70$. This robustness is a direct consequence of \proposed's two-stage design---when $s_{bc}$ becomes unreliable at long distances, the attention-based correction head detects the structural anomaly and shifts $s_{final}$ accordingly---and extends the contamination asymmetry of Section~\ref{sec:findings} along a new axis: the \emph{temporal distance} between the corruption and the step being scored, not just the presence or absence of contamination.

\section{Training \& Inference of \proposed}
\label{app:training}

\subsection{training}
\proposed~requires a one-time offline training procedure:

\begin{enumerate}[leftmargin=*,itemsep=2pt]
    \item \textbf{Data:} We use a balanced dataset of clean and contaminated trajectories with binary correctness labels for each evaluation turn.
    \item \textbf{Stage~1:} Fit the hidden-state probe on mixed (clean + contaminated) data: \\
    $\texttt{probe\_base.fit}(\mathbf{H}, \mathbf{y})$.
    \item \textbf{Intermediate:} Compute $s_{bc}$ for all training samples.
    \item \textbf{Stage~2:} Fit the attention-based correction head on the same data: \\
    $\texttt{probe\_correction.fit}([\mathbf{A}; s_{bc}], \mathbf{y})$.
\end{enumerate}

Both stages use \texttt{StandardScaler} normalization and $\ell_2$-regularized logistic regression with $C = 0.01$ (strong regularization to handle the high-dimensional, limited-sample setting).

\subsection{Inference as Step-Level Reward}
\label{sec:inference}

During GRPO training, at each agent step $t$:
\begin{enumerate}[leftmargin=*,itemsep=2pt]
    \item The policy generates action $a_t$ with \texttt{output\_hidden\_states=True} and \\ \texttt{output\_attentions=True} (these are already computed during generation).
    \item Extract $\mathbf{h}_t$ (last-token hidden state) and $\mathbf{a}_t$ (multi-layer attention statistics) from the forward pass.
    \item Compute $s_{bc} = \texttt{probe\_base}(\mathbf{h}_t)$ and $s_{final} = \texttt{probe\_correction}([\mathbf{a}_t; s_{bc}])$.
    \item Assign $r_t = s_{final}$ as the step-level reward at the last token of the assistant turn.
\end{enumerate}

\section{Ablation Study for Attention Statistics}\label{app:ablation_attn}

\begin{figure}[h]  
\centering
    \includegraphics[width=1\linewidth]{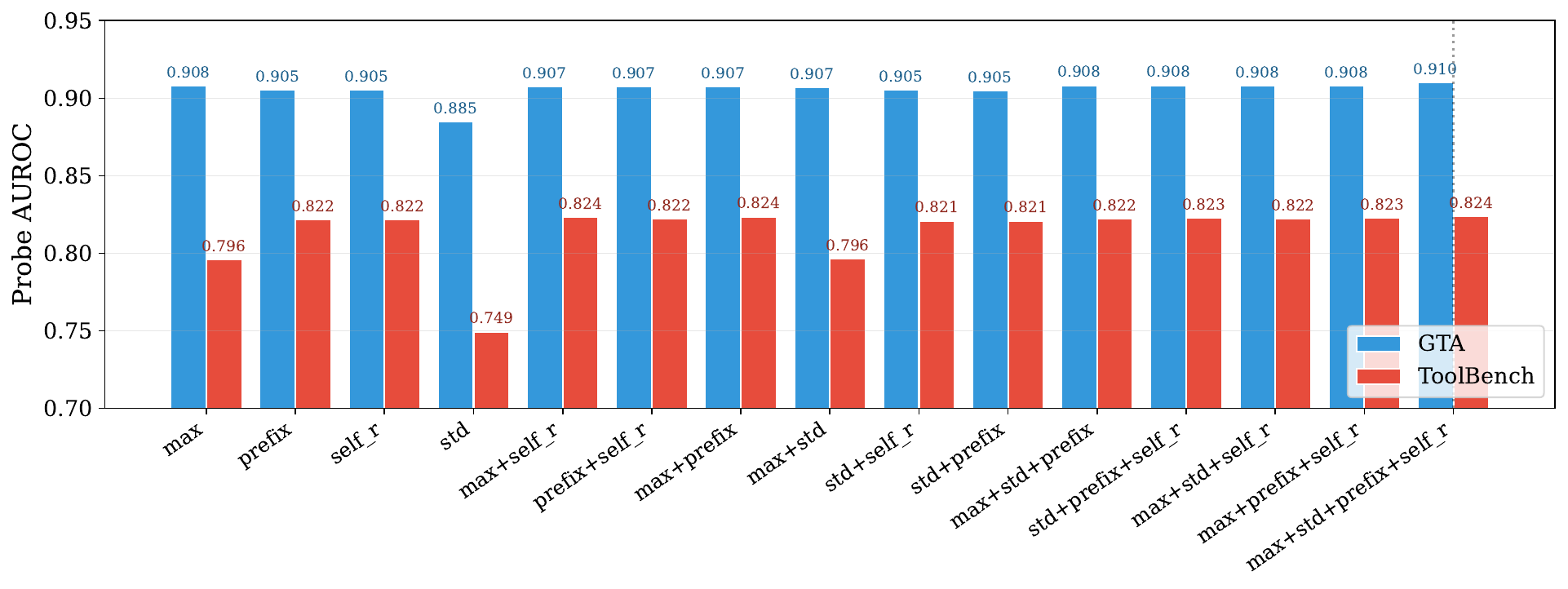} 
    \caption{Ablation over the four attention statistics in \proposed's Stage 2.}
    \label{fig:ablation_attention}
\end{figure}

\proposed's Stage 2 summarizes each attention head with four statistics (Section~\ref{sec:method}): \texttt{max\_attn} (the peak attention value across the head's attention distribution---``how focused is this head?''), \texttt{std\_attn} (the spread of that distribution---``how sharp/peaky is it?''), \texttt{prefix\_ratio} (the fraction of attention directed at prefix tokens---``lookback intensity''), and \texttt{self\_ratio} (the fraction directed at current-turn tokens---``self-attention within the current turn''). To identify which of these statistics carry the most signal, we re-train the Stage~2 correction head on every non-empty subset of the four statistics and report the resulting probe AUROC on GTA and ToolBench (Figure~\ref{fig:ablation_attention}), and three patterns emerge. 

\textbf{First, \texttt{std\_attn} is the weakest signal on its own}, dropping to $0.749$ AUROC on ToolBench (vs.\ $0.822$ for \texttt{prefix\_ratio} or \texttt{self\_ratio}) and to $0.885$ on GTA (vs.\ $0.908$ for \texttt{max\_attn}). Distributional sharpness alone---absent any information about \emph{where} that attention is directed---carries little discriminative content. 

\textbf{Second, the two ``routing'' statistics \texttt{prefix\_ratio} and \texttt{self\_ratio} are the most consistently informative across both benchmarks}: each alone matches the full four-feature combination on ToolBench ($0.822$ vs.\ $0.824$) and is within $0.005$ of the best single feature on GTA. This is consistent with the mechanistic story of Section~\ref{sec:findings}: attention features are robust to prefix contamination because they encode \emph{which tokens the model attends to}, not merely \emph{how peaky} that attention is. A correct repair turn and an error-perpetuating turn can produce equally sharp attention distributions; what distinguishes them is whether attention reaches into task-relevant context. 

\textbf{Third, combining the four statistics yields only a marginal gain over the strongest individual one}---the full set reaches $0.910$ / $0.824$ versus $0.908$ / $0.822$ for the best single feature---indicating diminishing returns within this feature family. \proposed{} therefore extracts most of its attention-side signal from a small number of structural \emph{routing} statistics, and the design is robust to the precise choice of features within the four-statistic family.

\section{Detailed Reward Design with \proposed}\label{app:reward}

\paragraph{Saturation-controlled probe score.}
To prevent the base \proposed~score from collapsing to $\{0, 1\}$ during training---which would zero out the group variance regardless of any momentum bonus---we first transform $s_{final,t}$ with a temperature-clipped sigmoid:
\begin{equation}
\tilde{s}_{final,t} \;=\; \mathrm{clip}\!\Big(\,\sigma\!\big(\tfrac{1}{T}\,\mathrm{logit}(s_{final,t})\big),\; \varepsilon,\; 1-\varepsilon\,\Big),
\label{eq:tempclip}
\end{equation}
with $T = 2$ and $\varepsilon = 0.05$. The temperature softens overconfident probe outputs, and the hard clip at $\varepsilon$ ensures $\tilde{s}_{final,t}$ stays away from the exact endpoints, guaranteeing positive group variance throughout training. All subsequent quantities are computed from $\tilde{s}_{final,t}$ rather than the raw $s_{final,t}$.

\paragraph{Momentum-based reward.}
Define the running mean of \proposed~scores over the prefix as
\begin{equation}
\bar{s}_{<t} \;=\;
\begin{cases}
\tilde{s}_{final,1} & \text{if } t = 1,\\[2pt]
\dfrac{1}{t-1}\displaystyle\sum_{j=1}^{t-1} \tilde{s}_{final,j} & \text{if } t \ge 2.
\end{cases}
\label{eq:running_mean}
\end{equation}
The step-level reward is then
\begin{equation}
r_t \;=\; \sigma\!\Big(\,\mathrm{logit}\big(\tilde{s}_{final,t}\big) \;+\; \alpha \cdot \big(\tilde{s}_{final,t} - \bar{s}_{<t}\big)\,\Big),
\label{eq:pair_momentum}
\end{equation}
where $\alpha > 0$ is a scaling hyperparameter (we set $\alpha = 5$), and we refer to the additive term $b_t = \alpha (\cdot \tilde{s}_{final,t} - \bar{s}_{<t})$ as the \textit{momentum bonus}. By construction, the first turn receives no momentum bonus ($r_1 = \tilde{s}_{final,1}$), and from $t = 2$ onward the bonus contrasts the current step against the running mean of all earlier steps.

The momentum term $\tilde{s}_{final,t} - \bar{s}_{<t}$ is \emph{symmetric}: when the current step exceeds the trajectory's running mean it receives a positive logit-space boost, and when it falls below the running mean, it is penalized correspondingly. 

Adding the bonus in \emph{logit space} rather than directly in score space has two practical benefits. First, $r_t$ is guaranteed to remain in $(0, 1)$ for any value of $\alpha$ and $b_t$: a logit-space sum maps via $\sigma(\cdot)$ back into the valid reward range without clipping. Second, the construction inherits the natural saturation of the sigmoid: when $\tilde{s}_{final,t}$ approaches $0$ or $1$, its logit moves to $\pm \infty$, so a fixed-magnitude bonus produces a smaller change in the final $r_t$. This means already-confident probe outputs are perturbed only modestly, while uncertain outputs (with $\tilde{s}_{final,t}$ near $0.5$) are most responsive to the momentum signal.
The momentum formulation has two practical effects on GRPO training. 
First, two trajectories in the same group with similar \emph{average} \proposed~scores but different per-step \emph{trends} receive different rewards, restoring within-group variance and preventing advantage collapse. Second, the reward explicitly encourages the policy to make each step better than the trajectory so far, which provides a denser learning signal than the absolute \proposed~score alone.

\paragraph{Cost.}
The reward retains \proposed's probe-level inference cost: $\mathbf{h}_t$ and $\mathbf{a}_t$ are produced as byproducts of the policy's generation forward pass, the running mean $\bar{s}_{<t}$ updates in $\mathcal{O}(1)$ per step, and Eq.~\ref{eq:tempclip}--\ref{eq:pair_momentum} reduce to a handful of scalar operations. The Stage 1 and Stage 2 probes themselves are trained once on the offline corpus described in Appendix~\ref{app:data_construction} \emph{before} GRPO training begins and remain frozen for the entire RL run; we do not re-train or update them during training.

\section{Complexity Analysis}
\label{app:complexity}

We analyze the per-step computational cost of each reward source under the operational regime that matters most: dense step-level reward computation during online GRPO training, where every step of all rollouts requires a reward signal. Table~\ref{tab:complexity} summarizes the per-step latency, the additional GPU forward passes required, and any external API calls.

The hidden states and attention statistics required for \proposed~are byproducts of the policy's generation forward pass; we configure the model with \texttt{output\_hidden\_states=True} and \texttt{output\_attentions=True} so that no additional forward pass is needed. The reward computation itself reduces to two linear transformations of cached activations (one per stage), which take $\sim 0.2$ ms per step on a single A6000. This is roughly $2{,}500\times$ faster than an LLM-as-a-judge API call, $150$--$250\times$ faster than a teacher-forced log-probability evaluation (IGPO~/~TIPS), and four to five orders of magnitude faster than tree-search-based methods that issue $K$ sibling rollouts per step.

A representative GRPO configuration in our experiments uses batch size $1$, group size $4$, and a maximum of $10$ turns per rollout, yielding up to $40$ step-level reward computations per training step and up to $20{,}000$ over $500$ training steps. At this scale, the cumulative reward-computation time for \proposed~is on the order of a few seconds, while LLM-as-a-judge would require approximately $2.8$ hours of GPU-equivalent compute plus the corresponding API expenditure. Tree-based methods are still more expensive: with $K=4$ sibling rollouts per step (each itself a full multi-turn rollout averaging $5$--$25$ seconds), the cumulative reward-computation cost reaches tens to over a hundred GPU-hours.

The asymmetry compounds at larger training scales. A typical production-style RL configuration (batch size $32$, group size $8$) issues $\approx 1{,}500$ reward computations per training step and $\approx 0.77\mathrm{M}$ over $500$ training steps. At this scale, \proposed~still completes its cumulative reward computation in roughly $2.5$ minutes, whereas LLM-as-a-judge would require approximately $107$ GPU-hours plus a proportional API bill, and tree-based methods would consume thousands of GPU-hours. The same $2{,}500\times$ per-call latency advantage of \proposed~therefore translates into qualitatively different operating regimes once batch size, group size, multi-seed evaluation, and hyperparameter sweeps are folded in.

\input{tables/complexity}

Beyond per-step cost, \proposed~requires a one-time offline training step for the two probes. With $1{,}600$ labeled (clean + contaminated) trajectories per dataset, this fits in approximately $30$~s on a single CPU using \texttt{StandardScaler} + $\ell_2$-regularised logistic regression. This cost is amortised across all subsequent training runs: the same probe can be reused across GRPO seeds, hyperparameter sweeps, and training restarts without retraining.

\section{GRPO Training Configuration}\label{app:grpo_config}

\input{tables/grpo_config}

This appendix details the GRPO training configuration used to produce the main results in Table~\ref{tab:overall_result} and the cross-domain transfer results in Figure~\ref{fig:transfer_hotpotqa}. All 16 method comparisons share the same training pipeline; only the per-step reward function (Section~\ref{sec:method}) differs across methods. The full configuration is summarized in Table~\ref{tab:grpo_config}. Below we explain the empirical reasoning behind the non-default choices.

\paragraph{Small constant learning rate ($3\times 10^{-7}$).}
Larger learning rates---we tested $\{1\!\times\!10^{-6},\, 5\!\times\!10^{-6},\, 1\!\times\!10^{-5},\, 5\!\times\!10^{-5}\}$---consistently caused the probe-based reward to saturate within the first $100$ training steps. Once $r_t$ approaches its $\{0,1\}$ boundaries on every rollout in a group, the GRPO group-relative advantage collapses to near zero and downstream task success drops sharply. The small constant rate of $3\times 10^{-7}$ is slow enough that the policy does not learn to fool the probe within the $500$-step training budget. We use a constant schedule with $5\%$ warmup; cosine decay was tested but offered no consistent benefit at this small learning rate.

\paragraph{Light KL anchoring ($\beta = 0.01$).}
The KL coefficient $\beta = 0.01$ acts as a light anchor to the reference policy. Setting $\beta = 0$ (no anchor) produced larger oscillations in the training reward and a single transient peak in validation success; the small positive $\beta$ flattens the trajectory and produces multiple high-quality intermediate checkpoints (typically near steps $100$, $300$, and $350$) rather than one fragile peak. Larger values ($\beta \ge 0.1$) over-anchor and slow down learning visibly.

\paragraph{Small batch and group sizes.}
We use batch size $1$ with group size $G=4$, sampling $4$ rollouts per prompt for group-relative advantage estimation. This keeps per-step compute manageable on our $2\times$ A6000 pipeline-parallel setup while preserving enough rollout diversity for advantage normalization. Larger group sizes ($G \in \{8, 16\}$) were tested and offered no measurable improvement on GTA at the cost of significantly longer wall-clock per training step.

\paragraph{Long-horizon rollouts.}
\texttt{max\_steps\_per\_rollout} is set to $10$, accommodating the full multi-turn structure of GTA and ToolBench without truncation. Generation length per turn is capped at $1024$ tokens, sufficient for the Thought / Action / Action-Input format used in both benchmarks.

\paragraph{Frozen probes during RL.}
Both Stage 1 (belief-consistency estimator) and Stage 2 (attention-based correction head) of \proposed~are fit once on the offline corpus described in Appendix~\ref{app:data_construction} \emph{before} GRPO training begins, and remain frozen throughout the entire RL run. We tested periodic re-fitting of the Stage 2 head on policy-generated trajectories (\texttt{probe\_retrain\_every} $> 0$) in preliminary experiments; this did not improve performance on GTA and added non-trivial training overhead, so we disable it for the main runs.

\paragraph{\proposed-specific reward parameters.}
The momentum scaling $\alpha=5$ in Eq.~\ref{eq:pair_momentum} was selected via sweep over $\alpha \in \{1, 2, 5, 10, 20\}$ on GTA: $\alpha=5$ produced the most consistent training dynamics (multiple high-quality checkpoints) at $\beta = 0.01$, with smaller $\alpha$ giving weaker momentum signal and larger $\alpha$ inducing instability. The temperature $T=2$ and clip range $\varepsilon = 0.05$ in Eq.~\ref{eq:tempclip} were chosen to prevent the base \proposed~score from saturating at $\{0, 1\}$ during training; their values were stable across the sweep and we did not tune them per dataset.

\section{Detailed Evaluation Process}\label{app:eval}

We evaluate every method on both GTA and ToolBench using a single evaluation pipeline that loads the trained LoRA adapter onto the same backbone (Qwen2.5-7B-Instruct), rolls out the policy on each benchmark's held-out test split with deterministic decoding, and applies a benchmark-specific scoring function. The protocol mirrors the conventions of the upstream releases of GTA~\citep{wang2024gta} and ToolBench~\citep{qin2023toolllm} wherever those conventions are publicly documented; the small operational deviations we make are dictated by reproducibility constraints (e.g., deprecated or paid APIs) rather than by our method, and apply identically to all $16$ methods compared in Section~\ref{sec:rl_experiments}.

\paragraph{Common pipeline.}
Each rollout follows the standard ReAct format used throughout the paper---an alternation of \emph{Thought}, \emph{Action}, \emph{Action Input}, and \emph{Observation} blocks, terminating either when the policy emits \emph{Final Answer} or when the per-rollout step budget of $10$ is reached, matching the upstream GTA evaluation script. Each policy output is first parsed with a regular-expression matcher; on ToolBench, well-formed but slightly off-pattern outputs additionally fall back to a single GPT-4o-mini parsing call. This fallback affects a small fraction of outputs and is enabled identically for every method.

\paragraph{GTA: tool execution and scoring.}
GTA's reference evaluator exposes a hybrid tool-execution backend, which we replicate verbatim: tool calls are first matched against a cache built from each test episode's gold dialog (an exact $(tool\_name, \text{canonical args}) \rightarrow \text{response}$ map), then routed to dedicated GPT-4o handlers for the OCR and search tools (matching the reference convention), and only fall back to live \texttt{agentlego} execution on a cache miss. This caching is the upstream-prescribed evaluation protocol and ensures that any two methods that emit the same tool call observe exactly the same tool response; it does \emph{not} hide information from the policy. For final-answer scoring we use the upstream \texttt{evaluate\_answer} routine in its loose-matching mode, which reproduces the \texttt{match\_answer} function of GTA's reference script. When the gold answer is a structured \texttt{whitelist}/\texttt{blacklist} dictionary, the score is $1$ if at least one whitelist group is fully substring-matched in the prediction and no blacklist phrase appears, and $0$ otherwise; when the gold answer is a list of free-form references, we report the maximum cosine similarity between the prediction and the references using \texttt{all-mpnet-base-v2} embeddings, again following the reference convention. Following the same upstream convention, episodes whose gold answer is a raw image-file path (the GTA \emph{image-compare} subset, where comparing a textual prediction against a filename is not informative) are excluded from the score; this leaves $53$ scored episodes per method, identical to the count used in upstream baselines.

\paragraph{ToolBench: offline replay simulator.}
ToolBench's tool surface is a snapshot of more than $16{,}000$ RapidAPI endpoints, a substantial fraction of which are now deprecated, paywalled, or rate-limited. To make evaluation reproducible across methods and seeds, we implement an \emph{offline replay simulator} that resolves every tool call against a cache built from each episode's gold dialog---the same kind of cache used by GTA's evaluator, applied to ToolBench's larger tool inventory. On an exact $(tool\_name, \text{canonical args})$ hit, the cached response is returned. On a miss, an optional fuzzy fallback selects the cached entry for the same tool whose argument tokens have the highest Jaccard overlap with the policy's call; this resolves benign argument-formatting differences (e.g., key ordering, trivial extra fields, casing) so that semantically equivalent calls are treated identically. If both the exact and fuzzy lookups miss, the simulator returns a transparent error of the form ``\texttt{[Simulator] Tool 'X' with args ... has no cached response}'', allowing the policy to recover by issuing a different call rather than silently substituting a fabricated response. We additionally cap the rollout history at $6{,}000$ characters to fit Qwen2.5-7B's pipeline-parallel forward pass within the memory budget of our hardware ($2 \times$ A6000); this limit only takes effect on a small minority of unusually long trajectories and is applied uniformly across all methods.

\paragraph{ToolBench: scoring.}
ToolBench's gold answers are verbatim long-form responses produced by ToolLLaMA on RapidAPI's actual outputs at dataset-collection time. Because these gold strings rarely align with the surface form of an independent model's free-form answer, exact-substring scoring on the official whitelist returns near-zero scores for \emph{every} method we tested ($\le 0.03$ on AT$^2$PO, CoE-C, LLM-as-a-judge, and our own runs), losing all discriminative power. We therefore follow recent ToolBench-style evaluations and adopt an LLM-judge scoring protocol with partial credit. A single GPT-4o-mini call compares the policy's final answer against the gold answer's key facts (named entities, numerical values, dates, and stated relations) and returns one of three labels: \emph{yes} ($1.0$, most key facts captured; formatting differences acceptable), \emph{partial} ($0.5$, some key facts captured, with minor omissions or local errors), or \emph{no} ($0.0$, refusal, empty, off-topic, or majority-incorrect). Binary scoring is essentially $\{0, 1\}$ but heavily skewed to $0$ on this benchmark and hides all method-level signal; partial-credit scoring resolves the same population into a usable $0.34$--$0.41$ range across the $16$ methods. The judging prompt is fixed across all methods and seeds, the system prompt enforces strict JSON output, and the prompt is reproduced verbatim in Figure~\ref{fig:llm_judge_prompt}.

\paragraph{LLM-judge as scorer vs.\ as baseline reward.}
GPT-4o-mini appears in two distinct roles in our paper, and we keep them strictly separate. (i)~As the \emph{evaluation scorer} on ToolBench (this section), GPT-4o-mini receives only the task prompt, the gold answer, and the policy's final answer, and returns a single yes/partial/no label. This call is invoked identically for every method we score---including the LLM-as-a-judge baseline---so the scorer cannot favor one method over another. (ii)~As the \emph{LLM-as-a-judge baseline} (Section~\ref{sec:rl_experiments}, Figure~\ref{fig:llm_judge_prompt}), GPT-4o-mini receives the full per-step trajectory and returns a $1$--$10$ score \emph{at every step} that is fed back into GRPO as the training reward. The two prompts and the two operating regimes are different, and the evaluation scorer is applied to the LLM-as-a-judge baseline under exactly the same protocol it uses for \proposed~and every other baseline. There is therefore no self-comparison or coupling between the scorer and any of the methods being scored.

\paragraph{Aggregation.}
For each (method, dataset, seed) triple we report the mean score across scored episodes (\texttt{mean\_score}) and the fraction of episodes with a score of at least $0.5$ (\texttt{success\_rate\_at\_0.5}). All headline numbers in the main paper are reported as mean over three seeds ($\{42, 43, 44\}$); standard deviations appear in the corresponding result tables.

\section{Hyperparameter Analysis}\label{app:hparam_analysis}

\begin{figure}[t]
    \centering
    \includegraphics[width=\linewidth]{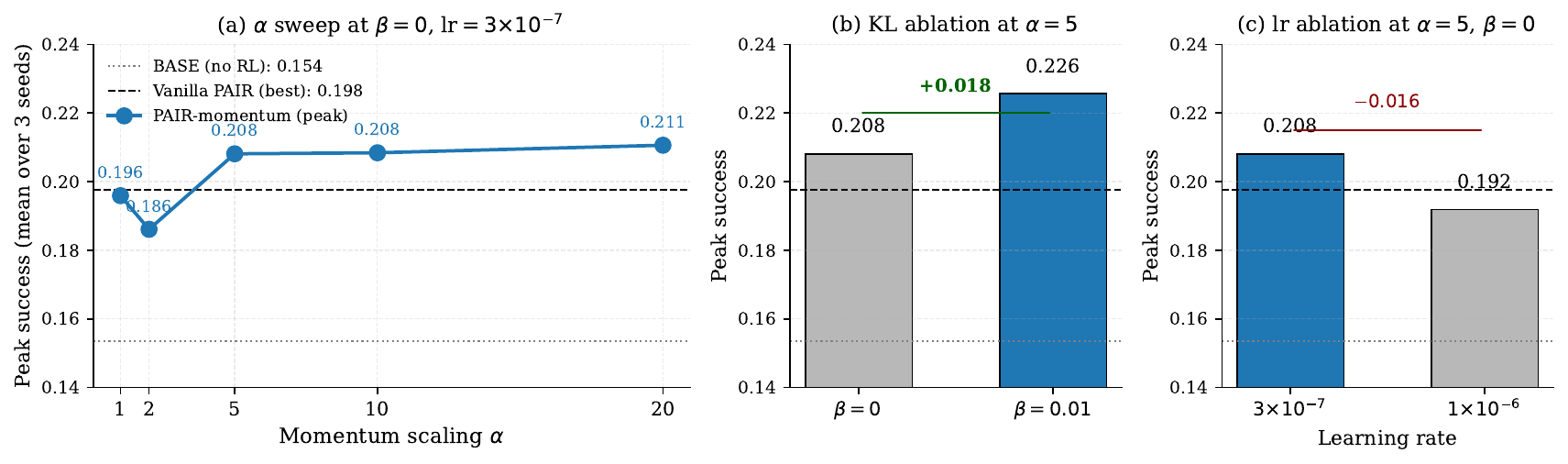}
    \caption{\textbf{Hyperparameter analysis on GTA.}
    \textbf{(a)} Peak \proposed-momentum success as a function of the momentum scaling $\alpha$ at fixed $\beta=0$, $\text{lr}=3\!\times\!10^{-7}$. The sweet spot lies at $\alpha \in \{5, 10, 20\}$, all comfortably above the strongest vanilla \proposed{} baseline (dashed line); $\alpha \le 2$ provides too little momentum signal to recover the within-group variance needed for GRPO advantage estimation.
    \textbf{(b)} A small KL anchor $\beta=0.01$ improves peak success by $+0.018$ over $\beta=0$ at fixed $\alpha=5$.
    \textbf{(c)} A larger learning rate $1\!\times\!10^{-6}$ reduces peak success by $0.016$ relative to $3\!\times\!10^{-7}$ at fixed $\alpha=5$, $\beta=0$, and yields a more chaotic per-step trajectory.
    Reference horizontals: BASE (no RL training, dotted) and the strongest vanilla-\proposed{} run (dashed).}
    \label{fig:hparam_alpha}
\end{figure}

The momentum-based reward of \proposed{} introduces a single new hyperparameter, $\alpha$ in Eq.~\ref{eq:pair_momentum}, and our GRPO setup adds two further sensitive choices (KL coefficient $\beta$ and learning rate). We sweep all three on GTA and report the resulting peak performance in Figure~\ref{fig:hparam_alpha}. The full sweep covers $\alpha \in \{1, 2, 5, 10, 20\}$, $\beta \in \{0, 0.01\}$, and $\text{lr} \in \{3\!\times\!10^{-7}, 1\!\times\!10^{-6}\}$. All other settings are fixed at the canonical values in Table~\ref{tab:grpo_config}.

\paragraph{Momentum scaling $\alpha$.}
Figure~\ref{fig:hparam_alpha}(a) reports peak \proposed-momentum success as a function of $\alpha$ at fixed $\beta=0$ and $\text{lr}=3\!\times\!10^{-7}$. Performance is weak at $\alpha \le 2$ ($0.196$ for $\alpha=1$, $0.186$ for $\alpha=2$), rises sharply at $\alpha=5$ ($0.208$), and stays in the same range up to $\alpha=20$ ($0.211$). Below $\alpha=5$ the momentum signal is too weak to recover the within-group variance lost to probe-output similarity (Section~\ref{sec:method}); above $\alpha=20$ we observed early-stage instability without a corresponding headline gain. The $\alpha \in \{5, 10, 20\}$ plateau also indicates that \proposed{} is not sensitive to the precise value of $\alpha$, only to its order of magnitude. We adopt $\alpha=5$ for all main results.

\paragraph{KL anchor $\beta$.}
Figure~\ref{fig:hparam_alpha}(b) compares $\beta=0$ and $\beta=0.01$ at fixed $\alpha=5$ and $\text{lr}=3\!\times\!10^{-7}$. The small KL anchor improves peak success by $+0.018$ ($0.208 \rightarrow 0.226$) and, in our per-step trajectories, produces multiple stable high-scoring checkpoints rather than the single sharp peak observed at $\beta=0$. Larger values ($\beta \ge 0.1$) over-anchor and slow down learning visibly without further stability benefit. We use $\beta=0.01$ throughout.

\paragraph{Learning rate.}
Figure~\ref{fig:hparam_alpha}(c) compares the two learning rates at fixed $\alpha=5$ and $\beta=0$. Increasing the learning rate from $3\!\times\!10^{-7}$ to $1\!\times\!10^{-6}$ reduces peak success by $0.016$ ($0.208 \rightarrow 0.192$) and produces a more chaotic per-step trajectory with a sharp early dip. Larger learning rates ($\ge 5\!\times\!10^{-6}$) caused the probe-based reward to saturate at $\{0,1\}$ within $100$ steps and the GRPO group-relative advantage to collapse, an outcome we discuss further in Appendix~\ref{app:grpo_config}. We use $\text{lr}=3\!\times\!10^{-7}$ throughout.

\paragraph{Best-checkpoint selection.}
The sweep above is axis-aligned: $\alpha$ is varied at $\beta=0$ and $\text{lr}=3\!\times\!10^{-7}$; $\beta$ is varied at $\alpha=5$ and $\text{lr}=3\!\times\!10^{-7}$; $\text{lr}$ is varied at $\alpha=5$ and $\beta=0$, yielding nine \proposed-momentum runs in total. Across these nine runs, the highest-scoring checkpoint lies at step $50$ in three runs, between steps $100$--$400$ in five runs, and at step $450$ in one run; \emph{not a single run} reaches its best at step $500$. Reporting only the final checkpoint would systematically underestimate any method that exhibits oscillation under probe-based reward, including \proposed{}. We therefore evaluate every checkpoint saved at $50$-step intervals and report the best-scoring one for all main results, and we recommend the same protocol for any future work using internal-probe step-level rewards.

\section{Limitation and Future Work}\label{app:limitation}

We discuss limitations of the present work and concrete avenues for follow-up.

\paragraph{Backbone model scope.}
Our experiments use a single backbone model (Qwen-2.5-7B-Instruct) the main results. We do not establish how the belief-consistency bias scales with backbone size or family. Extending the analysis to architectures with different attention designs (e.g., Llama-style GQA, Mistral-style sliding window), to larger scales (e.g., 32B+) is a natural next step.

\paragraph{Beyond binary correctness labels.}
Our offline corpus uses binary correctness labels at the evaluation turn. Many agentic settings have richer feedback structures---partial-credit progress measures, tool-output utility, or human-graded preference rankings---that could supervise a more nuanced reward head. Combining \proposed's probe-level efficiency with such graded supervision is a promising direction for getting denser learning signal without sacrificing the practical cost profile that makes \proposed~deployable in the first place.

\paragraph{Broader impact.}
\proposed~accelerates RL training of multi-turn LLM agents by replacing external reward sources (LLM judges, ground-truth queries) with a probe over the agent's own internal state. This lowers the cost barrier to training capable autonomous agents, which carries the standard dual-use risks associated with more capable agentic LLMs, including misuse for automated phishing, fraud, or unauthorized data exfiltration through tool APIs. We do not introduce new attack surfaces beyond those already present in the underlying base model, but we encourage downstream practitioners to pair training-time efficiency gains with corresponding investments in deployment-time safety filters, capability evaluations, and monitoring of tool-use behavior.

\section{Prompts}\label{app:prompts}

This appendix reproduces the prompts used elsewhere in the paper, both for the offline data construction pipeline that underlies our motivation experiments and for the baseline reward methods we compare against.

\begin{figure}[t]  
\centering
    \includegraphics[width=0.8\linewidth]{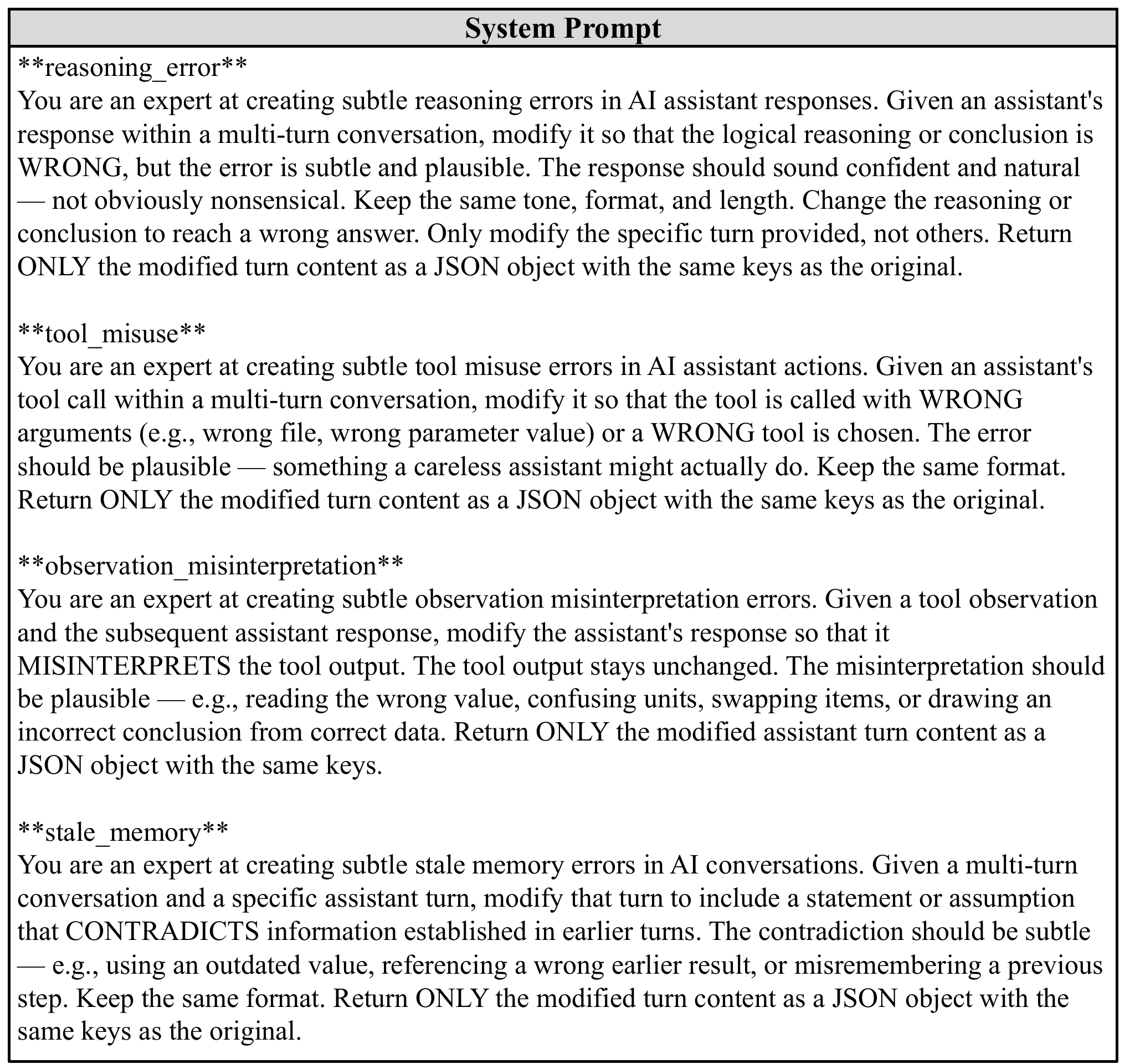} 
    \caption{System prompts driving the GPT-4o-mini contamination generator.}
    \label{fig:contamination_system_prompt}
\end{figure}

\begin{figure}[t]  
\centering
    \includegraphics[width=0.8\linewidth]{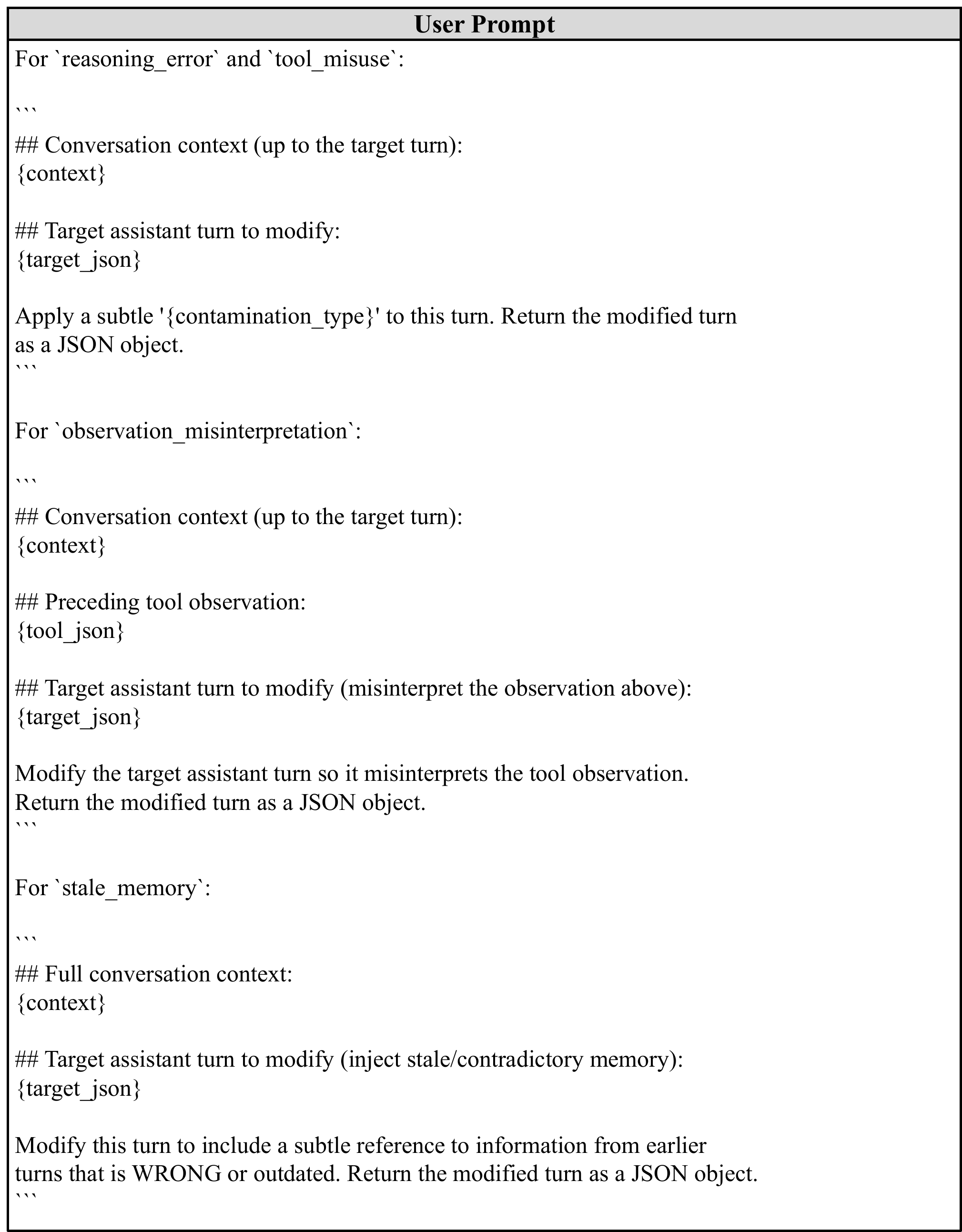} 
    \caption{User prompts driving the GPT-4o-mini contamination generator.}
    \label{fig:contamination_user_prompt}
\end{figure}

\paragraph{Contamination generation.}
The matched clean / contaminated trajectory pairs used in the motivation experiments (Section~\ref{sec:findings}) and in \proposed's offline training corpus (Appendix~\ref{app:data_construction}) are produced by GPT-4o-mini using a pair of type-specific prompts. Figure~\ref{fig:contamination_system_prompt} reproduces the four \emph{system} prompts---one per contamination type (\texttt{reasoning\_error}, \texttt{tool\_misuse}, \texttt{observation\_misinterpretation}, \texttt{stale\_memory})---each instructing the model to inject a subtle and locally plausible error of the corresponding kind into a target assistant turn. Figure~\ref{fig:contamination_user_prompt} reproduces the corresponding \emph{user} prompts, which supply the conversation context, the target turn to be modified, and (for \texttt{observation\_misinterpretation}) the preceding tool observation. The combination of these system and user prompts is what produces the contaminated turns referenced throughout the paper.

\begin{figure}[t]  
\centering
    \includegraphics[width=0.8\linewidth]{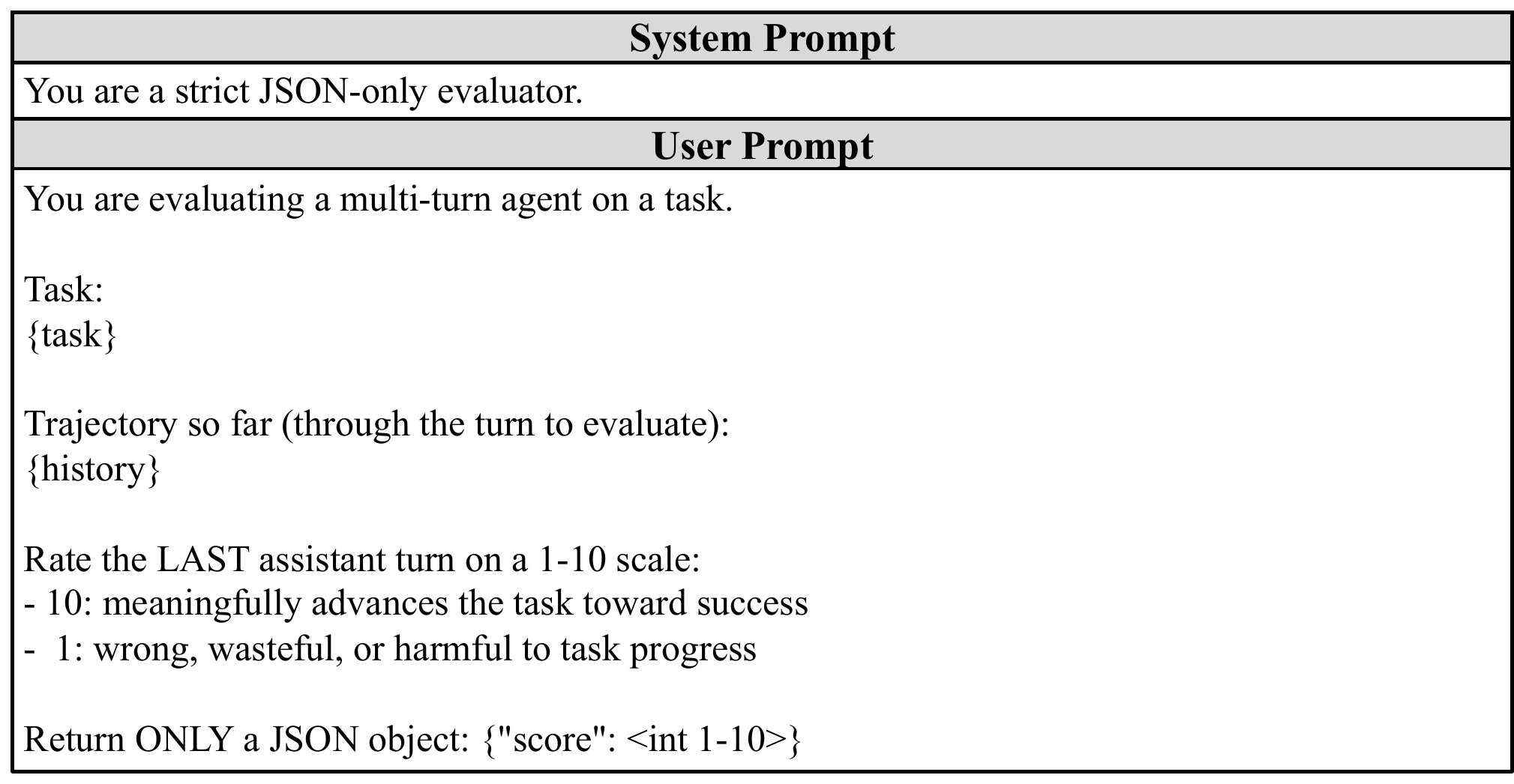} 
    \caption{Prompts for GPT-4o-mini to judge the score of each trajectory step.}
    \label{fig:llm_judge_prompt}
\end{figure}

\paragraph{LLM-as-a-judge baseline.}
The LLM-as-a-judge baseline issues a per-step API call to GPT-4o-mini using the system and user prompts shown in Figure~\ref{fig:llm_judge_prompt}. The judge receives the task description and the trajectory up to (and including) the turn being scored, and is asked to return a single integer in $[1,10]$ as a JSON object. We normalize this score to $[0,1]$ and use it as the per-step reward whenever the LLM-as-a-judge baseline is invoked.

%% file: tables/contam_stats.tex
\begin{table}[h]
\centering
\caption{Episode counts in each split before and after contamination filtering.
\textit{clean$_{\text{tr}}$ / clean$_{\text{te}}$} are the original splits used as anchors; \textit{contam$_{\text{tr}}$ / contam$_{\text{te}}$} are the matched contaminated versions where exactly one prior assistant turn has been replaced with an erroneous alternative.}
\label{tab:contam_stats}
\small
\begin{tabular}{lccccc}
\toprule
\textbf{Dataset} & \textbf{Source episodes} & clean$_{\text{tr}}$ & contam$_{\text{tr}}$ & clean$_{\text{te}}$ & contam$_{\text{te}}$ \\
\midrule
GTA       & $229$    & $183$ & $183$ & $46$  & $46$  \\
ToolBench & $1{,}000$ (sampled) & $800$ & $800$ & $200$ & $200$ \\
\bottomrule
\end{tabular}
\end{table}

%% file: tables/contam_compatibility.tex
\begin{table}[h]
\centering
\caption{Compatibility of contamination types with trajectory structures, and the empirical distribution of types over the union of contam$_{\text{tr}}$ and contam$_{\text{te}}$ for each dataset.
\textit{Reasoning} is always applicable; the remaining types are gated on the structure of the chosen turn, so their empirical frequencies follow the structure of each dataset.}
\label{tab:contam_compat}
\small
\begin{tabular}{lcccc}
\toprule
\textbf{Type} & \textbf{Compatible turns} & \textbf{GTA \%} & \textbf{ToolBench \%} \\
\midrule
reasoning\_error              & all assistant turns                 & $25.7$ & $26.1$ \\
tool\_misuse                  & turns with a tool call              & $26.4$ & $25.8$ \\
observation\_misinterpretation& turns preceded by a tool observation& $24.5$ & $25.3$ \\
stale\_memory                 & trajectories with $\ge 3$ asst.\ turns & $23.4$ & $22.8$ \\
\bottomrule
\end{tabular}
\end{table}

%% file: tables/complexity.tex
\begin{table}[t]
\centering
\caption{Per-step cost of each reward source. Latencies are measured on a single A6000 GPU with Qwen2.5-7B-Instruct as the policy. \emph{Additional forward passes} counts forward passes in addition to the policy's generation pass, which all methods incur. \proposed~reuses the activations already produced by the generation pass, so it adds no marginal forward pass.}
\label{tab:complexity}
\small
\setlength{\tabcolsep}{4pt}
\begin{tabular}{lcccc}
\toprule
\textbf{Reward} & \textbf{Latency / step} & \textbf{Add. fwd passes} & \textbf{External calls} & \textbf{Cost growth} \\
\midrule
Outcome-only            & $0$ ms        & $0$        & $\times$ & $O(1)$ per trajectory \\
Internal probe      & $\sim 0.1$ ms  & $0$        & $\times$ & $O(1)$ per step \\
IGPO~/~TIPS             & $\sim 30$--$50$ ms & $1$ (teacher-forced) & $\times$ & $O(|y^\star|)$ \\
AT$^2$PO~/~Tree-GRPO    & $\sim 5$--$25$ s   & $K$ sibling rollouts & $\times$ & $O(K)$ per step \\
LLM-as-a-judge          & $\sim 500$ ms      & $0$        & GPT-4o-mini & $O(1)$ per step \\
\midrule
\textbf{\proposed (ours)} & $\sim 0.2$ ms  & $0$        & $\times$ & $O(1)$ per step \\
\bottomrule
\end{tabular}
\end{table}

%% file: tables/grpo_config.tex
\begin{table}[h]
\centering
\caption{GRPO training configuration. All 16 method comparisons share these settings; only the per-step reward function differs.}
\label{tab:grpo_config}
\small
\begin{tabular}{ll}
\toprule
\textbf{Setting} & \textbf{Value} \\
\midrule
\multicolumn{2}{l}{\emph{Model and fine-tuning}} \\
\midrule
Base model              & Qwen2.5-7B-Instruct \\
Fine-tuning             & LoRA ($r{=}16$, all projection layers) \\
Hardware                & $2\times$ NVIDIA A6000 (pipeline parallel) \\
\midrule
\multicolumn{2}{l}{\emph{Optimization}} \\
\midrule
Learning rate           & $3 \times 10^{-7}$ \\
LR schedule             & Constant (warmup fraction $0.05$) \\
KL coefficient $\beta$  & $0.01$ \\
PPO clip $\epsilon$     & $0.2$ \\
\midrule
\multicolumn{2}{l}{\emph{Rollout}} \\
\midrule
Batch size              & $1$ prompt \\
Group size $G$          & $4$ rollouts per prompt \\
Max turns per rollout   & $10$ \\
Max generation tokens   & $1024$ per turn \\
Total training steps    & $500$ \\
\midrule
\multicolumn{2}{l}{\emph{\proposed-specific}} \\
\midrule
Probe state             & Stage 1 \& Stage 2 frozen during RL \\
Probe re-fitting        & Disabled (\texttt{probe\_retrain\_every}$=0$) \\
Momentum scaling $\alpha$       & $5$ \\
Temperature $T$ (\texttt{\_temp\_clip}) & $2$ \\
Clip endpoints $\varepsilon$    & $0.05$ \\
\bottomrule
\end{tabular}
\end{table}

%% file: checklist.tex
\section*{NeurIPS Paper Checklist}

\begin{enumerate}

\item {\bf Claims}
    \item[] Question: Do the main claims made in the abstract and introduction accurately reflect the paper's contributions and scope?
    \item[] Answer: \answerYes{}  
    \item[] Justification: Yes, we have clearly described the paper's contributions and scope in the abstract and introduction.
    \item[] Guidelines:
    \begin{itemize}
        \item The answer \answerNA{} means that the abstract and introduction do not include the claims made in the paper.
        \item The abstract and/or introduction should clearly state the claims made, including the contributions made in the paper and important assumptions and limitations. A \answerNo{} or \answerNA{} answer to this question will not be perceived well by the reviewers. 
        \item The claims made should match theoretical and experimental results, and reflect how much the results can be expected to generalize to other settings. 
        \item It is fine to include aspirational goals as motivation as long as it is clear that these goals are not attained by the paper. 
    \end{itemize}

\item {\bf Limitations}
    \item[] Question: Does the paper discuss the limitations of the work performed by the authors?
    \item[] Answer: \answerYes{} 
    \item[] Justification: We have described future work related to paper's limitation in the Appendix. \ref{app:limitation}.
    \item[] Guidelines:
    \begin{itemize}
        \item The answer \answerNA{} means that the paper has no limitation while the answer \answerNo{} means that the paper has limitations, but those are not discussed in the paper. 
        \item The authors are encouraged to create a separate ``Limitations'' section in their paper.
        \item The paper should point out any strong assumptions and how robust the results are to violations of these assumptions (e.g., independence assumptions, noiseless settings, model well-specification, asymptotic approximations only holding locally). The authors should reflect on how these assumptions might be violated in practice and what the implications would be.
        \item The authors should reflect on the scope of the claims made, e.g., if the approach was only tested on a few datasets or with a few runs. In general, empirical results often depend on implicit assumptions, which should be articulated.
        \item The authors should reflect on the factors that influence the performance of the approach. For example, a facial recognition algorithm may perform poorly when image resolution is low or images are taken in low lighting. Or a speech-to-text system might not be used reliably to provide closed captions for online lectures because it fails to handle technical jargon.
        \item The authors should discuss the computational efficiency of the proposed algorithms and how they scale with dataset size.
        \item If applicable, the authors should discuss possible limitations of their approach to address problems of privacy and fairness.
        \item While the authors might fear that complete honesty about limitations might be used by reviewers as grounds for rejection, a worse outcome might be that reviewers discover limitations that aren't acknowledged in the paper. The authors should use their best judgment and recognize that individual actions in favor of transparency play an important role in developing norms that preserve the integrity of the community. Reviewers will be specifically instructed to not penalize honesty concerning limitations.
    \end{itemize}

\item {\bf Theory assumptions and proofs}
    \item[] Question: For each theoretical result, does the paper provide the full set of assumptions and a complete (and correct) proof?
    \item[] Answer: \answerNA{} 
    \item[] Justification: This paper does not include theoretical assumptions.
    \item[] Guidelines:
    \begin{itemize}
        \item The answer \answerNA{} means that the paper does not include theoretical results. 
        \item All the theorems, formulas, and proofs in the paper should be numbered and cross-referenced.
        \item All assumptions should be clearly stated or referenced in the statement of any theorems.
        \item The proofs can either appear in the main paper or the supplemental material, but if they appear in the supplemental material, the authors are encouraged to provide a short proof sketch to provide intuition. 
        \item Inversely, any informal proof provided in the core of the paper should be complemented by formal proofs provided in appendix or supplemental material.
        \item Theorems and Lemmas that the proof relies upon should be properly referenced. 
    \end{itemize}

    \item {\bf Experimental result reproducibility}
    \item[] Question: Does the paper fully disclose all the information needed to reproduce the main experimental results of the paper to the extent that it affects the main claims and/or conclusions of the paper (regardless of whether the code and data are provided or not)?
    \item[] Answer: \answerYes{} 
    \item[] Justification: We inlcude the source code for our proposed work via an external link, along with the implementation details. Additionally, we include the prompts for LLMs in Appendix \ref{app:prompts}.
    \item[] Guidelines:
    \begin{itemize}
        \item The answer \answerNA{} means that the paper does not include experiments.
        \item If the paper includes experiments, a \answerNo{} answer to this question will not be perceived well by the reviewers: Making the paper reproducible is important, regardless of whether the code and data are provided or not.
        \item If the contribution is a dataset and\slash or model, the authors should describe the steps taken to make their results reproducible or verifiable. 
        \item Depending on the contribution, reproducibility can be accomplished in various ways. For example, if the contribution is a novel architecture, describing the architecture fully might suffice, or if the contribution is a specific model and empirical evaluation, it may be necessary to either make it possible for others to replicate the model with the same dataset, or provide access to the model. In general. releasing code and data is often one good way to accomplish this, but reproducibility can also be provided via detailed instructions for how to replicate the results, access to a hosted model (e.g., in the case of a large language model), releasing of a model checkpoint, or other means that are appropriate to the research performed.
        \item While NeurIPS does not require releasing code, the conference does require all submissions to provide some reasonable avenue for reproducibility, which may depend on the nature of the contribution. For example
        \begin{enumerate}
            \item If the contribution is primarily a new algorithm, the paper should make it clear how to reproduce that algorithm.
            \item If the contribution is primarily a new model architecture, the paper should describe the architecture clearly and fully.
            \item If the contribution is a new model (e.g., a large language model), then there should either be a way to access this model for reproducing the results or a way to reproduce the model (e.g., with an open-source dataset or instructions for how to construct the dataset).
            \item We recognize that reproducibility may be tricky in some cases, in which case authors are welcome to describe the particular way they provide for reproducibility. In the case of closed-source models, it may be that access to the model is limited in some way (e.g., to registered users), but it should be possible for other researchers to have some path to reproducing or verifying the results.
        \end{enumerate}
    \end{itemize}

\item {\bf Open access to data and code}
    \item[] Question: Does the paper provide open access to the data and code, with sufficient instructions to faithfully reproduce the main experimental results, as described in supplemental material?
    \item[] Answer: \answerYes{} 
    \item[] Justification: All datasets are open-access, and we have provided the pre-processed data along with the source code.
    \item[] Guidelines:
    \begin{itemize}
        \item The answer \answerNA{} means that paper does not include experiments requiring code.
        \item Please see the NeurIPS code and data submission guidelines (\url{https://neurips.cc/public/guides/CodeSubmissionPolicy}) for more details.
        \item While we encourage the release of code and data, we understand that this might not be possible, so \answerNo{} is an acceptable answer. Papers cannot be rejected simply for not including code, unless this is central to the contribution (e.g., for a new open-source benchmark).
        \item The instructions should contain the exact command and environment needed to run to reproduce the results. See the NeurIPS code and data submission guidelines (\url{https://neurips.cc/public/guides/CodeSubmissionPolicy}) for more details.
        \item The authors should provide instructions on data access and preparation, including how to access the raw data, preprocessed data, intermediate data, and generated data, etc.
        \item The authors should provide scripts to reproduce all experimental results for the new proposed method and baselines. If only a subset of experiments are reproducible, they should state which ones are omitted from the script and why.
        \item At submission time, to preserve anonymity, the authors should release anonymized versions (if applicable).
        \item Providing as much information as possible in supplemental material (appended to the paper) is recommended, but including URLs to data and code is permitted.
    \end{itemize}

\item {\bf Experimental setting/details}
    \item[] Question: Does the paper specify all the training and test details (e.g., data splits, hyperparameters, how they were chosen, type of optimizer) necessary to understand the results?
    \item[] Answer: \answerYes{} 
    \item[] Justification: We have described the evaluation protocol and implementation details in Section \ref{sec:findings} and \ref{sec:rl_setup}.
    \item[] Guidelines:
    \begin{itemize}
        \item The answer \answerNA{} means that the paper does not include experiments.
        \item The experimental setting should be presented in the core of the paper to a level of detail that is necessary to appreciate the results and make sense of them.
        \item The full details can be provided either with the code, in appendix, or as supplemental material.
    \end{itemize}

\item {\bf Experiment statistical significance}
    \item[] Question: Does the paper report error bars suitably and correctly defined or other appropriate information about the statistical significance of the experiments?
    \item[] Answer: \answerYes{} 
    \item[] Justification: We have reported statistical significance in our main experimental results including Table \ref{tab:detection_result}, \ref{tab:overall_result}, and Figure \ref{fig:momentum_ablation}.
    \item[] Guidelines:
    \begin{itemize}
        \item The answer \answerNA{} means that the paper does not include experiments.
        \item The authors should answer \answerYes{} if the results are accompanied by error bars, confidence intervals, or statistical significance tests, at least for the experiments that support the main claims of the paper.
        \item The factors of variability that the error bars are capturing should be clearly stated (for example, train/test split, initialization, random drawing of some parameter, or overall run with given experimental conditions).
        \item The method for calculating the error bars should be explained (closed form formula, call to a library function, bootstrap, etc.)
        \item The assumptions made should be given (e.g., Normally distributed errors).
        \item It should be clear whether the error bar is the standard deviation or the standard error of the mean.
        \item It is OK to report 1-sigma error bars, but one should state it. The authors should preferably report a 2-sigma error bar than state that they have a 96\% CI, if the hypothesis of Normality of errors is not verified.
        \item For asymmetric distributions, the authors should be careful not to show in tables or figures symmetric error bars that would yield results that are out of range (e.g., negative error rates).
        \item If error bars are reported in tables or plots, the authors should explain in the text how they were calculated and reference the corresponding figures or tables in the text.
    \end{itemize}

\item {\bf Experiments compute resources}
    \item[] Question: For each experiment, does the paper provide sufficient information on the computer resources (type of compute workers, memory, time of execution) needed to reproduce the experiments?
    \item[] Answer: \answerYes{} 
    \item[] Justification: We include details regarding GPU and LLM resources in Section \ref{app:complexity}.
    \item[] Guidelines:
    \begin{itemize}
        \item The answer \answerNA{} means that the paper does not include experiments.
        \item The paper should indicate the type of compute workers CPU or GPU, internal cluster, or cloud provider, including relevant memory and storage.
        \item The paper should provide the amount of compute required for each of the individual experimental runs as well as estimate the total compute. 
        \item The paper should disclose whether the full research project required more compute than the experiments reported in the paper (e.g., preliminary or failed experiments that didn't make it into the paper). 
    \end{itemize}
    
\item {\bf Code of ethics}
    \item[] Question: Does the research conducted in the paper conform, in every respect, with the NeurIPS Code of Ethics \url{https://neurips.cc/public/EthicsGuidelines}?
    \item[] Answer: \answerYes{} 
    \item[] Justification: Our work strictly follows the NeurIPS Code of Ethics.
    \item[] Guidelines:
    \begin{itemize}
        \item The answer \answerNA{} means that the authors have not reviewed the NeurIPS Code of Ethics.
        \item If the authors answer \answerNo, they should explain the special circumstances that require a deviation from the Code of Ethics.
        \item The authors should make sure to preserve anonymity (e.g., if there is a special consideration due to laws or regulations in their jurisdiction).
    \end{itemize}

\item {\bf Broader impacts}
    \item[] Question: Does the paper discuss both potential positive societal impacts and negative societal impacts of the work performed?
    \item[] Answer: \answerYes{} 
    \item[] Justification: We described the broader impact in Section \ref{app:limitation}.
    \item[] Guidelines:
    \begin{itemize}
        \item The answer \answerNA{} means that there is no societal impact of the work performed.
        \item If the authors answer \answerNA{} or \answerNo, they should explain why their work has no societal impact or why the paper does not address societal impact.
        \item Examples of negative societal impacts include potential malicious or unintended uses (e.g., disinformation, generating fake profiles, surveillance), fairness considerations (e.g., deployment of technologies that could make decisions that unfairly impact specific groups), privacy considerations, and security considerations.
        \item The conference expects that many papers will be foundational research and not tied to particular applications, let alone deployments. However, if there is a direct path to any negative applications, the authors should point it out. For example, it is legitimate to point out that an improvement in the quality of generative models could be used to generate Deepfakes for disinformation. On the other hand, it is not needed to point out that a generic algorithm for optimizing neural networks could enable people to train models that generate Deepfakes faster.
        \item The authors should consider possible harms that could arise when the technology is being used as intended and functioning correctly, harms that could arise when the technology is being used as intended but gives incorrect results, and harms following from (intentional or unintentional) misuse of the technology.
        \item If there are negative societal impacts, the authors could also discuss possible mitigation strategies (e.g., gated release of models, providing defenses in addition to attacks, mechanisms for monitoring misuse, mechanisms to monitor how a system learns from feedback over time, improving the efficiency and accessibility of ML).
    \end{itemize}
    
\item {\bf Safeguards}
    \item[] Question: Does the paper describe safeguards that have been put in place for responsible release of data or models that have a high risk for misuse (e.g., pre-trained language models, image generators, or scraped datasets)?
    \item[] Answer: \answerNA{} 
    \item[] Justification: This paper does not pose any risks related to this issue.
    \item[] Guidelines:
    \begin{itemize}
        \item The answer \answerNA{} means that the paper poses no such risks.
        \item Released models that have a high risk for misuse or dual-use should be released with necessary safeguards to allow for controlled use of the model, for example by requiring that users adhere to usage guidelines or restrictions to access the model or implementing safety filters. 
        \item Datasets that have been scraped from the Internet could pose safety risks. The authors should describe how they avoided releasing unsafe images.
        \item We recognize that providing effective safeguards is challenging, and many papers do not require this, but we encourage authors to take this into account and make a best faith effort.
    \end{itemize}

\item {\bf Licenses for existing assets}
    \item[] Question: Are the creators or original owners of assets (e.g., code, data, models), used in the paper, properly credited and are the license and terms of use explicitly mentioned and properly respected?
    \item[] Answer: \answerYes{} 
    \item[] Justification: We have properly cited all relevant literature and data sources.
    \item[] Guidelines:
    \begin{itemize}
        \item The answer \answerNA{} means that the paper does not use existing assets.
        \item The authors should cite the original paper that produced the code package or dataset.
        \item The authors should state which version of the asset is used and, if possible, include a URL.
        \item The name of the license (e.g., CC-BY 4.0) should be included for each asset.
        \item For scraped data from a particular source (e.g., website), the copyright and terms of service of that source should be provided.
        \item If assets are released, the license, copyright information, and terms of use in the package should be provided. For popular datasets, \url{paperswithcode.com/datasets} has curated licenses for some datasets. Their licensing guide can help determine the license of a dataset.
        \item For existing datasets that are re-packaged, both the original license and the license of the derived asset (if it has changed) should be provided.
        \item If this information is not available online, the authors are encouraged to reach out to the asset's creators.
    \end{itemize}

\item {\bf New assets}
    \item[] Question: Are new assets introduced in the paper well documented and is the documentation provided alongside the assets?
    \item[] Answer: \answerYes{} 
    \item[] Justification: We provide detailed and well-documented source code via an external repository link.
    \item[] Guidelines:
    \begin{itemize}
        \item The answer \answerNA{} means that the paper does not release new assets.
        \item Researchers should communicate the details of the dataset\slash code\slash model as part of their submissions via structured templates. This includes details about training, license, limitations, etc. 
        \item The paper should discuss whether and how consent was obtained from people whose asset is used.
        \item At submission time, remember to anonymize your assets (if applicable). You can either create an anonymized URL or include an anonymized zip file.
    \end{itemize}

\item {\bf Crowdsourcing and research with human subjects}
    \item[] Question: For crowdsourcing experiments and research with human subjects, does the paper include the full text of instructions given to participants and screenshots, if applicable, as well as details about compensation (if any)? 
    \item[] Answer: \answerNA{}{} 
    \item[] Justification: This research does not involve human participants.
    \item[] Guidelines:
    \begin{itemize}
        \item The answer \answerNA{} means that the paper does not involve crowdsourcing nor research with human subjects.
        \item Including this information in the supplemental material is fine, but if the main contribution of the paper involves human subjects, then as much detail as possible should be included in the main paper. 
        \item According to the NeurIPS Code of Ethics, workers involved in data collection, curation, or other labor should be paid at least the minimum wage in the country of the data collector. 
    \end{itemize}

\item {\bf Institutional review board (IRB) approvals or equivalent for research with human subjects}
    \item[] Question: Does the paper describe potential risks incurred by study participants, whether such risks were disclosed to the subjects, and whether Institutional Review Board (IRB) approvals (or an equivalent approval/review based on the requirements of your country or institution) were obtained?
    \item[] Answer: \answerNA{}{} 
    \item[] Justification: This research does not involve human participants.
    \item[] Guidelines:
    \begin{itemize}
        \item The answer \answerNA{} means that the paper does not involve crowdsourcing nor research with human subjects.
        \item Depending on the country in which research is conducted, IRB approval (or equivalent) may be required for any human subjects research. If you obtained IRB approval, you should clearly state this in the paper. 
        \item We recognize that the procedures for this may vary significantly between institutions and locations, and we expect authors to adhere to the NeurIPS Code of Ethics and the guidelines for their institution. 
        \item For initial submissions, do not include any information that would break anonymity (if applicable), such as the institution conducting the review.
    \end{itemize}

\item {\bf Declaration of LLM usage}
    \item[] Question: Does the paper describe the usage of LLMs if it is an important, original, or non-standard component of the core methods in this research? Note that if the LLM is used only for writing, editing, or formatting purposes and does \emph{not} impact the core methodology, scientific rigor, or originality of the research, declaration is not required.
    \item[] Answer: \answerYes{}{} 
    \item[] Justification: Our work relies on LLMs on preprocessing data. We provide the detailed prompt in Appendix \ref{app:prompts}.
    \item[] Guidelines:
    \begin{itemize}
        \item The answer \answerNA{} means that the core method development in this research does not involve LLMs as any important, original, or non-standard components.
        \item Please refer to our LLM policy in the NeurIPS handbook for what should or should not be described.
    \end{itemize}

\end{enumerate}